\documentclass[journal]{IEEEtran}
\ifCLASSINFOpdf
\else
\fi
\usepackage{times}
\usepackage{graphicx}
\usepackage{latexsym}

\usepackage{amsmath}
\usepackage{amsfonts}
\usepackage{multirow}
\usepackage{booktabs}
\usepackage{subfigure}

\DeclareMathOperator*{\softmax}{softmax}
\DeclareMathOperator*{\relu}{ReLU}

\def\W{\mathbf{W}}
\def\R{\mathbb{R}}
\def\L{\mathcal{L}}
\def\X{\mathbf{X}}

\def\P{\mathbf{P}}
\def\Q{\mathbf{Q}}
\def\w{\mathbf{w}}

\newcommand{\tabincell}[2]{\begin{tabular}{@{}#1@{}}#2\end{tabular}}

\begin{document}
%
\title{Risk Prediction on Traffic Accidents using a Compact Neural Model for Multimodal Information Fusion over Urban Big Data}
%
%
%

\author{Wenshan~Wang, 
        Su~Yang 
        and~Weishan~Zhang
\thanks{W. Wang was with Shanghai Key Laboratory of Intelligent Information Processing, School of
Computer Science, Fudan University, Shanghai, China}
\thanks{S. Yang was with Shanghai Key Laboratory of Intelligent Information Processing, School of
Computer Science, Fudan University, Shanghai, China e-mail: suyang@fudan.edu.cn.}
\thanks{W. Zhang was with College of Computer Science and Technology, China University of Petroleum, Qingdao, China.}
\thanks{Manuscript received April 00, 0000; revised August 00, 0000.}}

%
%

\markboth{IEEE Transactions,~Vol.~00, No.~0, August~0000}%
{Shell \MakeLowercase{\textit{et al.}}: Bare Demo of IEEEtran.cls for IEEE Journals}
%



\maketitle

\begin{abstract}
  Predicting risk map of traffic accidents is vital for accident prevention and early planning of emergency response. Here, the challenge lies in the multimodal nature of urban big data. We propose a compact neural ensemble model to alleviate overfitting in fusing multimodal features and develop some new features such as fractal measure of road complexity in satellite images, taxi flows, POIs, and road width and connectivity in OpenStreetMap. The solution is more promising in performance than the baseline methods and the single-modality data based solutions. After visualization from a micro view, the visual patterns of the scenes related to high and low risk are revealed, providing lessons for future road design. From city point of view, the predicted risk map is close to the ground truth, and can act as the base in optimizing spatial configuration of resources for emergency response, and alarming signs. To the best of our knowledge, it is the first work to fuse visual and spatio-temporal features in traffic accident prediction while advances to bridge the gap between data mining based urban computing and computer vision based urban perception.
\end{abstract}


%
\IEEEpeerreviewmaketitle

\section{INTRODUCTION}

Thanks to the availability of urban big data, cross-domain data analysis has been becoming increasingly important in the field of urban computing and urban perception. In the context of urban computing \cite{Zheng:2014:UCC:2648782.2629592}, some data sources can be leveraged for a variety of  prediction tasks, such as GPS trajectories \cite{Chen:2018:RRO:3178157.3161159}, Point-of-Interest (POI), human check-ins in location-based social networks \cite{Karamshuk:2013:GMO:2487575.2487616}, cell phone metadata, environmental data, etc. On the other hand, an emerging trend is the vision-based urban perception in the recent urban computing studies, which aims to percept social attributes in an urban region. In general, Google street view images or satellite images are used for urban perception. In the literature, several attempts \cite{Naik7571,Gebru13108} have been made to mine visual elements from Google street view images so as to predict non-visual city attributes, such as safety \cite{Porzi:2015:PUU:2733373.2806273}, housing prices, population density \cite{6875954}, city function, happiness \cite{Quercia:2014:ACM:2531602.2531613}, and liveness and depression \cite{10.1007/978-3-319-46448-0_12,DeNadai:2016:SLN:2964284.2964312}. Another trend is to utilize satellite imagery data so as to estimate non-visual social attributes \cite{Jean790}, which is a natural manner to observe an urban region globally. 

Recent evidence suggests that combining both non-visual urban data and visual data offers an effective way to make full use of complementary information from different data sources, which is a new trend in the cross-domain analysis. Although several attempts have made to combine satellite imagery data and urban big data including taxi trajectories and POI so as to predict commercial hotness \cite{He:2018:MCH:3301777.3287046}, it has rarely been investigated in the context of predicting traffic accident risk. Traffic accidents are a major public safety problem as traffic accidents often cause traffic jams and even cause loss of lives every year. According to the World Health Organization, it is reported that the number of annual road traffic deaths has reached 1.35 million in 2018 and road traffic injuries are the leading killer of people aged 5-29 years. Thus, inferring traffic accident risk has been becoming extremely important in the face of the high price paid for mobility. A map to indicate traffic accident risk is capable of providing the basis for optimizing the allocation of the resources for emergency response as well as traffic management. For example, we should set up more alarming tips in such places where traffic accidents are likely to occur.

In the literature, early works formulate traffic accident prediction as a classification or a regression problem by mapping features to the historical data of traffic accidents. The existing research works for traffic accident prediction are developed around the following clues: 1) A considerable amount of literature \cite{CHANG2005541,LIN2015444,CHANG2005365,CALIENDO2007657,EISENBERG2004637} is focused on utilizing machine learning models (e.g., Artificial Neural Network, Random Forests) based on environmental attributes such as weather \cite{BERGELHAYAT2013456}, road conditions so as to predict traffic accidents. 2) With the advance of deep learning techniques, recent works \cite{Chen:2016:LDR:3015812.3015863,Yuan:2018:HDL:3219819.3219922} propose to analyze traffic accidents based on deep models including Convolutional Neural Network (CNN), yielding superior performance.

So far, the previous studies \cite{Chen:2016:LDR:3015812.3015863,Najjar:2017:CSI:3298023.3298224} are focused on inferring traffic accident risk either based on spatial-temporal data or based on visual data, (e.g., satellite imagery). Despite the recent progress towards traffic accident risk prediction, few works have been done to combine non-visual data (e.g., spatial-temporal data) and visual data in the context of predicting traffic accident risk. The recent work \cite{He:2018:MCH:3301777.3287046} on commercial hotness prediction suggests that it is very promising to combine visual and non-visual data in prediction tasks in urban computing.

A series of surveys have been conducted to understand what features may impact traffic accident risk. In terms of non-visual features, previous studies \cite{Chen:2016:LDR:3015812.3015863,8397033} have reported that human mobility patterns, POI distribution, and road network patterns are discriminative features in predicting traffic accident risk. Intuitively, human mobility patterns extracted from taxi trajectories data or cell phone data can reflect traffic volume, road condition, and the complexity of road network to some extent. POI distribution in an urban region can indicate the city function \cite{Yang:2017:PCA:3139486.3130983}. Road network patterns directly indicate the complexity of road network. For visual features in predicting traffic accident risk, convolutional neural network representation achieves superior performance in predicting traffic accident risk in the recent work \cite{Najjar:2017:CSI:3298023.3298224}. Aside from the above features, in this paper, we claim that OpenStreetMap data can be used to extract discriminative features, such as connectivity of nodes in OpenStreetMap that indicates the complexity of road network, and road width feature. Additionally, we propose to extract the fractal feature from satellite images, which can figure out the complexity of the object of interest in a geometric sense from an overall point of view.

In this paper, we present a multimodal information fusion framework with dynamic weighting adaptive to the context to infer traffic accident risk. The objective of this research is to combine the spatio-temporal data and the satellite imagery data to infer the risk and find out the causes of traffic accidents. Specifically, we investigate a feature-level neural model, which consists of a classification network and a parameter prediction network. The key idea is that the parameter prediction network takes the data of a single-modality as input and then generates dynamic weights for the classification neural network to classify the data of another modality. The advantage of such a scheme is that the data of different modalities are deeply coupled at the feature weighting level. Furthermore, in \cite{He:2018:MCH:3301777.3287046}, non-visual data and visual data are incorporated into a context-aware neural network ensemble method to predict commercial hotness, which is a model-level fusion method. Inspired by that, we propose an improved method, yielding better performance in inferring traffic accident risk. Specifically, we utilize the parameter prediction network to generate dynamic weights for context-aware neural network, which is based on matrix decomposition to render compact representation of parameters against overfitting. The advantage of such a scheme is that decomposing parameter matrix into two low-rank matrices can reduce the number of parameters so as to alleviate overfitting, the nature of which is similar to Singular-Value Decomposition.

In order to find out the causes of traffic accidents, we present an elaborate visualization of the results of our models to get an intuitive understanding on satellite images. To be specific, the integrated gradients method \cite{Sundararajan:2017:AAD:3305890.3306024} is employed to visualize the cues that related to the high or low traffic accident risk, which is helpful for urban planning, traffic management, and emergency response.

In practice, we formulate the problem of predicting traffic accident risk as a three-category prediction problem: High, medium, and low. We conduct extensive experiments and the results show that our models achieve superior performance (The accuracy reaches 83\%.), outperforming the baseline models. Furthermore, our models demonstrate good interpretability, which is able to find out the causes of traffic accidents as indicated by the visual patterns of specific regions of interest in satellite images. The main contributions of our work can be summarized as follows:
\begin{itemize}
  \item To the best of our knowledge, combining the spatio-temporal urban big data and the satellite imagery data to infer traffic accident risk has rarely been investigated.
  \item We introduce a variety of features to infer traffic accident risk, including the road features obtained from OpenStreetMap, and the generalized fractal dimensions to measure the complexity of the region of interest on satellite images. To the best of our knowledge, these features have not been investigated in this task.
  \item We develop a multimodal fusion framework with dynamic weighting for inferring traffic accident risk. We present two multimodal dynamic fusion neural network at feature level and model level. We relax the context-aware neural network ensemble to a more compact representation by using matrix decomposition to obtain low-rank matrix represented parameters such that the overfitting problem can be alleviated.
  \item We show the interpretability of our models and the analyses allow us to find out the causes of traffic accidents based on the visualized patterns of satellite images.
\end{itemize}

\section{RELATED WORK}

A considerable amount of literature has been published in predicting traffic accident risk. In general, these studies consider it as a classification problem or a regression problem. Most of them are focused on investigating machine learning models, including Artificial Neural Network \cite{CHANG2005541}, and Decision Tree \cite{LIN2015444,ABELLAN20136047}. The previous works generally extract limited features from small scale training data, and then machine learning models are simply employed. More recently, attention has been being focused on utilizing deep learning models to analyze traffic accidents. Chen et al. \cite{Chen:2016:LDR:3015812.3015863} propose to make use of the stack denoise autoencoder model to learn representations from cell phone mobility data and attempt to predict real-time traffic accident risk, in reference to historical traffic accident records. Afterwards, Ren et al. \cite{DBLP:journals/corr/abs-1710-09543} attempt to predict traffic risk in the near future based on the long-short term memory (LSTM) model. Furthermore, Yuan et al. \cite{Yuan:2018:HDL:3219819.3219922} propose the Hetero-Conv-LSTM framework based on Convolutional LSTM so as to achieve the goal of real-time traffic accident prediction. However, more recent attention is focused on leveraging the open data, such as satellite imagery data, to assisting decision making for city-planning and policy formulation. Najjar et al. \cite{Najjar:2017:CSI:3298023.3298224} combine the satellite imagery data and historical accident data and takes them into convolutional neural network in order to predict traffic accident risk map. Our work is closer in spirit to \cite{Najjar:2017:CSI:3298023.3298224}. However, we deemed that it is not enough to employ single-modality data such as satellite imagery data. We reconsider the problem of inferring accident risk as a multimodal feature (including the spatio-temporal feature and visual feature) fusion problem. Furthermore, we present a method to interpret the results of our models based on satellite imagery data to understand the causes of traffic accidents.

\section{DYNAMIC INFORMATION FUSION FRAMEWORK FOR INFERRING TRAFFIC ACCIDENT RISK}
\begin{figure}[t]
  \centering
  \includegraphics[width=0.85\linewidth]{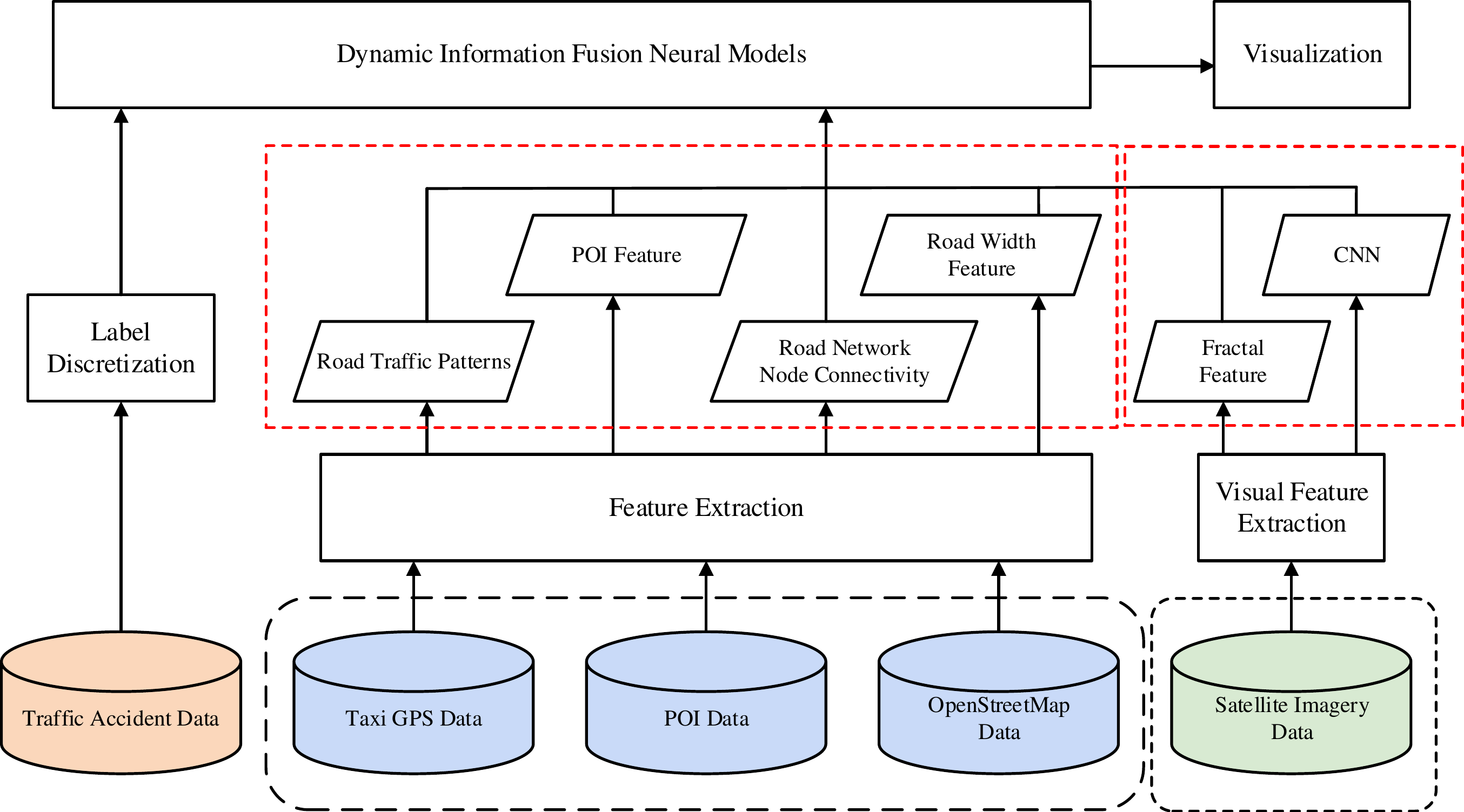}
  \caption{Dynamic information fusion framework for inferring traffic accident risk.}
  \label{fg:framework}
\end{figure}

\subsection{Preliminaries}
The main purpose of this investigation is to present an information fusion framework for multimodal data, which fuses satellite images and spatio-temporal urban big data in an adaptive manner to estimate the risk levels of traffic accidents all over the city to render a risk map as the visual guideline when preparing emergency response plans. Specifically, we consider the traffic accident risk inference problem as a classification problem, which can be formulated as follows: First, the geo-spatial area of interest is divided into grids and each grid $r$ represents a 1km $\times$ 1km sub-area, which is a proper area for traffic accident analysis in terms of size. Suppose that we have a set of grids $R = \left\{ {{r_1},{r_2}, \ldots ,{r_N}} \right\}$ consisting of $N$ sub-areas. Let $\mathbf{X}_u^i \in \mathbb{\R}^{D_u}$ and $\mathbf{X}_v^i \in \mathbb{R}^{D_v}$ denote the $D_u$-dimensional spatio-temporal feature and the $D_v$-dimensional visual feature of the $i$-th sub-area, respectively. Furthermore, we derive the traffic accident risk level $y^i$ of the $i$-th sub-area. Our objective is to learn a neural model to infer the traffic accident level ${{\hat y}^i} = f\left( { {\mathbf{X}_u^i, \mathbf{X}_v^i} } \right)$ based on the spatio-temporal feature and the visual feature. 

Figure \ref{fg:framework} presents an overview of the framework. First, the spatio-temporal feature is extracted from multiple data sources including Taxi GPS data, POI data, and OpenStreetMap data. Afterwards, the Bag-of-Words (BOW)\footnote{https://en.wikipedia.org/wiki/Bag-of-words\_model} method is employed to form the spatio-temporal feature: Road traffic patterns, POI feature, road network node connectivity, and road width feature. Meanwhile, the fractal feature and CNN representation are extracted from satellite imagery data to form the visual feature. Finally, the spatio-temporal feature and the visual feature are combined to be fed into the multimodal dynamic fusion neural models to infer the traffic accident risk level. Furthermore, a visualization method is present to find out possible causes of traffic accidents.

\subsection{Feature Extraction}
In this section, we describe the feature extraction methods as well as the motivation below. 

\paragraph{Road Traffic Patterns} 
Road traffic patterns from Taxi GPS data show a high correlation to traffic accident risk since traffic patterns are capable of disclosing the complexity of the topology of road network as well as traffic volume. Intuitively, road network with higher complexity in terms of topology has a higher risk of leading to a traffic accident. Another motivation to utilize road traffic patterns is that the risk of a traffic accident in a given area is in general relatively lower if there are less vehicles in this area.

Specifically, we define $I_t^i$ and $O_t^i$ as the number of taxis entering the $i$-th area and that coming out of it in time window $t$, respectively. We utilize an hour as the time interval. Then, we concatenate them as road traffic patterns:
\begin{align}
  \label{eq:tra}
  \mathbf{x}_{tra}^i = \left[ I_t^i, O_t^i  \right] 
\end{align}
where $\mathbf{x}_{tra}^i \in \mathbb{R}^{D_{tra}}$ is the $D_{tra}$-dimensional traffic pattern vector and $t = 1, 2, \ldots, 24$.


\paragraph{Point-of-Interest Feature}
The previous studies \cite{Yuan:2012:DRD:2339530.2339561,Yang:2017:PCA:3139486.3130983} have suggested that the distribution of POI categories in an area can characterize the urban functions affecting commercial activeness and act as the clue to predict the traffic volume regarding this area. Our intuition is that if the traffic volume in a commercial area is high as indicated by the POIs the traffic accident risk should correspondingly be high.

\begin{table}
  \center
  \caption{Each road width level contains the primary types of ``highway''.}
  \label{tab:width}
  \resizebox{0.42\textwidth}{!}{
  \begin{tabular}{cl}
    \toprule
    Width levels      &  Primary types of ``highway''  \\
    \midrule
    1    &  ``track'', ``living\_street'', ``crossing'', etc. \\
    2    &  ``servcie'', ``residential'', ``motorway\_junction'', etc.      \\
    3    &  ``secondary'', ``primary\_link'', ``tertiary'', etc.      \\
    4    &   ``motorway'', ``trunk'', etc. \\
    \bottomrule
  \end{tabular}
  }
\end{table}

Based on the above observation, we make use of the BOW method to quantify the POI distribution over all categories. Specifically, we count the number of POIs of each category in a region, which can be formulated as follows:
\begin{align}
   \label{eq:poi}
   \mathbf{x}_{poi}^i = {\left( {P_1^i,P_2^i, \ldots ,P_{D_{poi}}^i} \right)^T}
\end{align}
where $\mathbf{x}_{poi}^i \in \mathbb{R}^{D_{poi}}$ denotes the $D_{poi}$-dimensional vector of POI feature in the $i$-th area. That is, we consider $D_{poi}$ categories of POI in total. 

\paragraph{Road Network Node Connectivity}
The interconnected paved roads can form a road network. The complexity of road network is an important indicator of transportation. Intuitively, a more complicated road network leads to usually higher risk of traffic accidents.

To characterize the complexity of road network, we extract the feature from OpenStreetMap. OpenStreetMap\footnote{https://www.openstreetmap.org} is a free editable map of the world, which consists of nodes, ways, and relations\footnote{https://wiki.openstreetmap.org/wiki/Elements}. To be specific, we define the connectivity $N_j$ of the $j$-th node, indicating the number of the nodes connected to the $j$-th node. In the experiments, we found that $N_j$ obeys a long tail distribution. Most of nodes have low connectivity and only a small number of nodes have high connectivity. In practice, we bin $N_j$ into three complexity levels (high, medium, and low), and utilize the BoW method to quantify the distribution as follows:
\begin{align}
  \label{eq:con}
  \mathbf{x}_{con}^i = {\left( {CON_{high}^i, CON_{med}^i, CON_{low}^i} \right)^T}
\end{align}
where $\mathbf{x}_{con}^i \in \mathbb{R}^{D_{con}}$ is the $D_{con}$-dimensional node connectivity vector. $CON_{high}^i$, $CON_{med}^i$, and $CON_{low}^i$ denote the number of the nodes with high, medium, and low complexity levels in the $i$-th area, respectively.

\paragraph{Road Width Feature}

The width of roads has a high correlation to traffic accident risk as it is a key factor to cause traffic jams, which is a dangerous scenario to incur traffic accidents. It is generally agreed that a wider road corresponds with lower risk of traffic accidents.

In this paper, we also extract the road width feature from OpenStreetMap. According to the tag ``highway'' of the element in OpenStreetMap\footnote{https://wiki.openstreetmap.org/wiki/Map\_Features\#Highway}, we classify different types of ``highway'' into four levels. The fourth level is the widest one while the first level is the narrowest one. Each level contains the primary types of ``highway'', which are shown in Table \ref{tab:width}. Afterwards, the BoW method is employed to obtain the road width feature as follows:
\begin{align}
  \label{eq:wid}
  \mathbf{x}_{wid}^i = {\left( {WID_1^i, WID_2^i, WID_3^i, WID_4^i} \right)^T}
\end{align}
where $ \mathbf{x}_{wid}^i \in \mathbb{R}^{D_{wid}}$ is the $D_{wid}$-dimensional road width feature vector. $WID_1^i$, $WID_2^i$, $WID_3^i$, and $WID_4^i$ represent the number of different road width levels in the $i$-th region, respectively.

\paragraph{Fractal Feature}


\begin{figure}[t]
    \centering
    \subfigure[]{
    \label{fig:edge_1}
    \begin{minipage}[b]{0.95\linewidth}
    \includegraphics[width=0.46\linewidth]{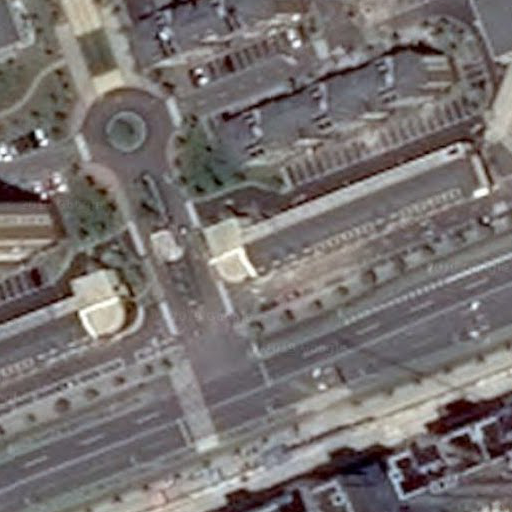}\hfill
    \includegraphics[width=0.46\linewidth]{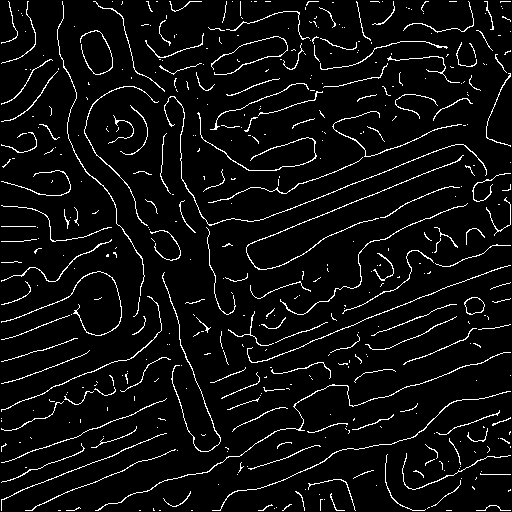}
    \end{minipage}
    }
    \hfill
    \subfigure[]{
    \label{fig:edge_2}
    \begin{minipage}[b]{0.95\linewidth}
    \includegraphics[width=0.46\linewidth]{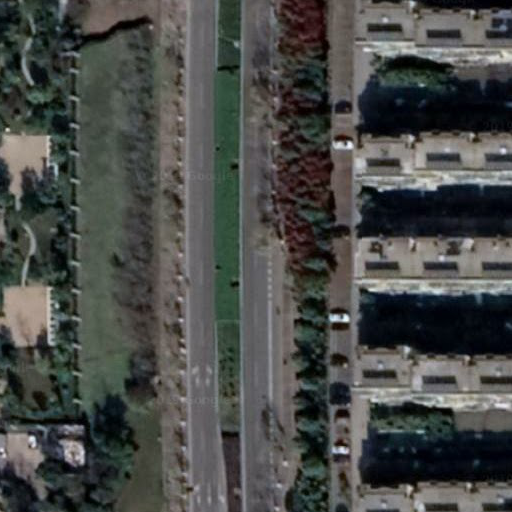}\hfill
    \includegraphics[width=0.46\linewidth]{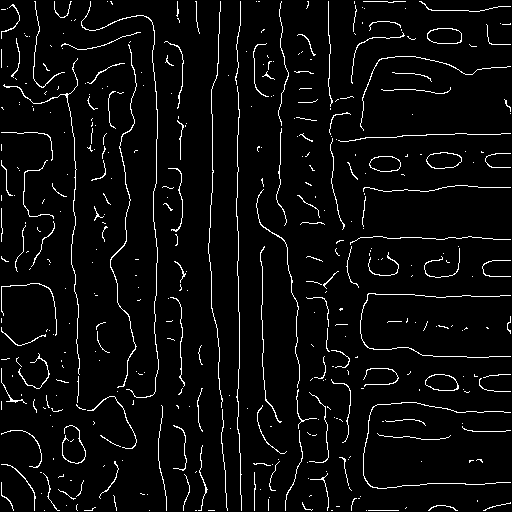}
    \end{minipage}
    }
    \caption{Original satellite images and the corresponding Canny edge maps.}
    \label{fg:edge}
\end{figure}

A fractal dimension figures out the complexity of an object\footnote{https://en.wikipedia.org/wiki/Fractal\_dimension}. The previous study \cite{4767591} has reported that the fractal dimension is capable of discriminating natural contexts from man-made objects. Inspired by that, we employ fractal dimension to characterize the surface complexity of the scenes of satellite images, which is a natural manner to observe a city from a global point of view and reveals richness and commercial activeness across the regions of interest \cite{jean2016combining,Wang:2018:UPC:3184558.3186581}.

In order to obtain the fractal feature, we make use of multifractal spectra \cite{5533703} based on the box-counting method. Specifically, we first randomly extract image patches with the size of 256 pixels $\times$ 256 pixels. We utilize the Canny edge detector \cite{4767851} to extract the edge map of each image patch. Intuitively, the edge map can capture the structure of the road network, which is illustrated in Figure \ref{fg:edge}. Then, the multifractal spectra method is employed to get a fractal feature vector for each edge map. Next, we utilize the K-means algorithm to cluster all the fractal feature vectors so as to build a visual dictionary. Then, vector quantization is applied to assign the fractal feature vectors of each satellite image to the dictionary to obtain a $D_{fra}$-dimensional BOW vector $\mathbf{x}_{fra}^m \in \mathbb{R}^{D_{fra}}$ for each sub-region.

\paragraph{Convolutional Neural Network}
Recently, Convolutional Neural Network (CNN) gains significant popularity for learning image representations in the computer vision community. Here, the reasons for the use of CNN are as follows: 1) CNN image representations are effective to figure out high-level objects in images. 2) We further visualize and interpret our inference model by using the technique based on CNN.

To be specific, ResNet-152 \cite{7780459} is employed as our CNN model, which is fine-tuned on satellite images. For an image $I$, it is vectorized into a $D_{cnn}$-dimensional feature vector $\mathbf{x}_{cnn}^i \in \mathbb{R}^{D_{cnn}}$ by the last layer of ResNet-152 as follows:
\begin{align}
  \label{eq:cnn}
  \mathbf{x}_{cnn}^i = {\rm{ResNet}}\left( I \right)
\end{align}
where ${\rm{ResNet}}\left(  \cdot  \right)$ denotes the ResNet-152 model.

\subsection{Traffic Accident Risk Level}
Assume that traffic accidents happen $n$ times at the $i$-th sub-area, the traffic accident risk level $y^i$ can be defined as follows:
\begin{align}
  \label{eq:accident}
  N_a^i = \sum\limits_{j = 1}^n {A_j^i}
\end{align}
where $A_j^i$ denotes the severity of $j$-th traffic accident event and $N_a^i$ is the sum of the severity at the $i$-th sub-area. In general, we let $A_j^i = 1$ when unknowing the severity.

Following \cite{Najjar:2017:CSI:3298023.3298224}, we make use of the K-means method \cite{Hartigan1979} to obtain three levels of traffic accident risk: High, medium, and low. That is ${y^i} \in \left\{ {0,1,2} \right\}$.

\subsection{Neural Models for Multimodal Information Fusion}
\label{sec:dynamic}

Following the aforementioned steps, we obtain the spatio-temporal feature $\mathbf{X}_u^i \in \mathbb{R}^{D_u}$ and visual feature $\mathbf{X}_v^i \in \mathbb{R}^{D_v}$, respectively, by using the concatenation operation $[\cdot]$:
\begin{align}
   \label{eq:combine}
   \mathbf{X}_u^i = [\mathbf{x}_{tra}^i, \mathbf{x}_{poi}^i, \mathbf{x}_{con}^i, \mathbf{x}_{wid}^i] \\
   \mathbf{X}_v^i = [\mathbf{x}_{fra}^i, \mathbf{x}_{cnn}^i]
\end{align}
where $D_u = D_{tra} + D_{poi} + D_{con} + D_{wid}$ and $D_v = D_{fra} + D_{cnn}$. Then, we can further get the multimodal feature vector $\mathbf{X}^i \in \mathbb{R}^{D_{u} + D_{v}}$ as follows:

\begin{align}
    \label{eq:multimodal}
    \mathbf{X}^i = [\mathbf{X}_u^i, \mathbf{X}_v^i]
\end{align}

Our goal is to learn the traffic accident risk inference model given the dataset \begin{math}\mathcal{D} = \{(\mathbf{X}^1, y^1), (\mathbf{X}^2, y^2), \ldots, (\mathbf{X}^N, y^N)\}\end{math}, which consists of $N$ regions. Except for the feature-level fusion as indicated in Eq. (\ref{eq:multimodal}), in this study, we develop the Feature-level Dynamic Fusion Neural Network (Feature-DFNN) and the Model-level Dynamic Fusion Neural Network (Model-DFNN).

\subsubsection{Feature-DFNN}

\begin{figure}[t]
  \centering
  \includegraphics[width=0.9\linewidth]{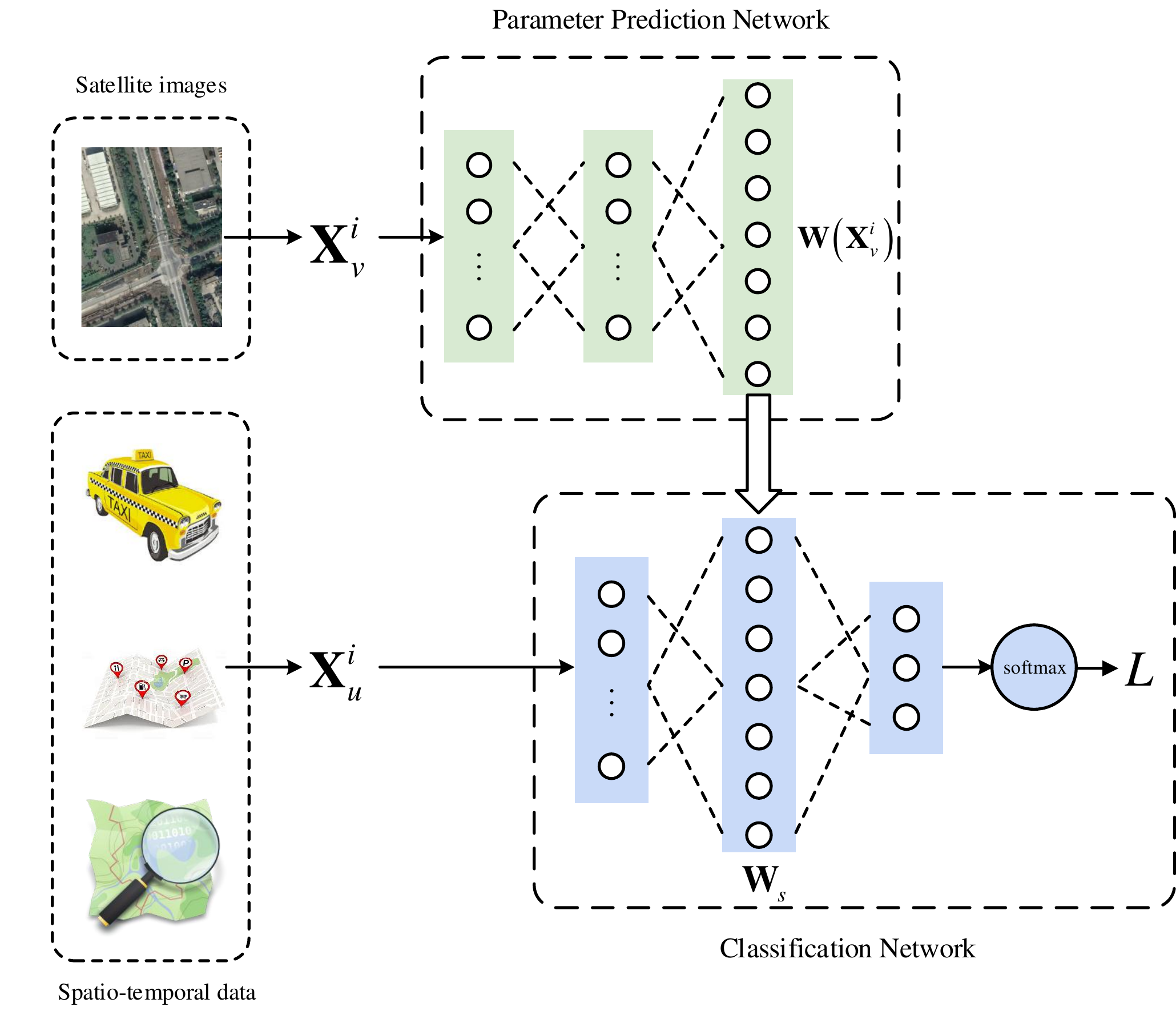}
  \caption{Feature-level DFNN, which consists of classification network and parameter prediction network. The parameter prediction network takes the visual feature as input and generates the dynamic parameters for the classification network to classify the spatio-temporal feature.}
  \label{fg:feature_dfnn}
\end{figure}

Feature-DFNN is composed of classification network and parameter prediction network. The basic idea is that the parameter prediction network takes the visual feature as input and generates the dynamic parameters for the classification network to classify the spatio-temporal feature so as to fuse multimodal modality sufficiently. Feature-DFNN is shown in Figure \ref{fg:feature_dfnn} and the details are described below:

\paragraph{Parameter Prediction Network}
The main purpose of the Parameter Prediction Network (PPNet) is to generate dynamic parameters subject to the visual feature on the fly for the classification network. It is challenging since predicting dynamic weights for another network is prone to overfitting. Taking inspiration from \cite{45823,NIPS2016_6068}, we apply the matrix decomposition technique to reduce the number of parameters, which is similar to Singular-Value Decomposition.

The parameter prediction network takes the visual feature $\X_v^i \in \R^{D_v}$ as input and then generates the dynamic weights $\W(\X_v^i)$, which can be formulated as follows: 

\begin{align}
    \label{eq:ppnet}
    \W(\X_v^i)&= {\rm {PPNet}}(\X_v^i) \\
              &= \P {\rm {Diag}}(\relu (\W_z \X_v^i) ) \Q
\end{align}
where $\rm Diag(\cdot)$ denotes the diagonal matrix function and $\relu(\cdot)$ denotes the non-linear function. $\W_z \in \R^{D_z \times D_v}$, $\P \in \R^{d \times D_z}$, and $\Q \in \R^{D_z \times d}$ denote the learnable parameters.



\paragraph{Classification Network}
The classification network aims to predict the traffic accident risk level, taking the spatio-temporal feature as input. The spatio-temporal feature $\mathbf{X}_u^i \in \mathbb{R}^{D_u}$ is passed into fully-connected layers followed by a softmax function.
\begin{align}
    \label{eq:dynamic}
    \hat {y}^i = \softmax( \W_2 \relu(\W_s \relu(\W_1 \mathbf{X}_u^i)))
\end{align}
where $\W_1 \in \R^{D_1 \times D_u}$, $\W_s \in \R^{D_s \times D_1}$, and $\W_2 \in \R^{D_2 \times D_s}$ are the parameters to be learned.

In order to achieve multimodal information fusion, we replace the static weights $\W_s$ in Eq. (\ref{eq:dynamic}) with the dynamic weights $\W(\X_v^i)$ in Eq. (\ref{eq:ppnet}) as follows:
\begin{align}
    \label{eq:together}
    \hat {y}^i &= \softmax( \W_2 \relu(\W(\X_v^i) \relu(\W_1 \mathbf{X}_u^i))) \\
               &= \softmax( \W_2 \relu({\rm {PPNet}}(\X_v^i) \relu(\W_1 \mathbf{X}_u^i)))
\end{align}

Here, the goal is to learn the parameters of the neural network as defined in Eq. (15) in an end-to-end manner by minimizing the cross-entropy loss defined as follows:
\begin{align}
    \label{eq:crossentropy}
    L(y, \hat{y}, \theta) =  - \frac{1}{N}\sum\limits_{i = 1}^N {\sum\limits_{c = 1}^C {y_c^i\log \left( {\hat y_c^i} \right)} }
\end{align}
where $C$ is the number of the traffic accident risk levels and $\theta$ is the weights set to be learned.

\subsubsection{Model-DFNN}
He et al. \cite{He:2018:MCH:3301777.3287046} propose a Context-aware Neural Network Ensemble (CNNE) method for predicting commercial hotness based on satellite images and social context data, which is a model-level fusion neural network for regression. In contrast to Feature-DFNN above, we found that there is a commonplace between them, that is,  generating dynamic weights for another network. The difference is that the dynamic parameters of Feature-DFNN relies on input only while the scheme of Model-DFNN is subject to both input and the corresponding output so as to remember the input-response behavior of each predictor and recall the confidence of such decision behavior in similar scenarios to obtain context-adaptive weighting of the decision.

In this paper, we investigate CNNE by incorporating the matrix decomposition technique for the traffic accident risk classification. By using the matrix decomposition based simplification of parameters, possible overfitting can be alleviated. More specifically, we first input the spatio-temporal feature $\X_u^i \in \R^{D_u}$ and the visual feature $\X_v^i \in \R^{D_v}$ into two classification neural networks that consist of two fully-connected layers, respectively. That is,
\begin{align}
    \label{eq:emsemble}
    \hat y_u^i = \softmax (\W_2^{(u)} \relu(\W_1^{(u)} \X_u^i)) \\
    \hat y_v^i = \softmax (\W_2^{(v)} \relu(\W_1^{(v)} \X_v^i) )
\end{align}
where $\W_1^{(u)} \in \R^{D_1^{(u)} \times D_u}$, $\W_2^{(u)} \in \R^{D_2^{(u)} \times D_1^{(u)}}$, $\W_1^{(v)} \in \R^{D_1^{(v)} \times D_v}$, and $\W_2^{(v)} \in \R^{D_2^{(v)} \times D_1^{(v)}}$ are the parameters to be learned. Then, the spatio-temporal feature and the visual feature are concatenated with the corresponding outputs of the classification neural networks, respectively, and as follows:
\begin{align}
    \label{eq:cat_out}
    \hat \X_u^i = [\X_u^i, \hat y_u^i] \\
    \hat \X_v^i = [\X_v^i, \hat y_v^i]
\end{align}
In \cite{He:2018:MCH:3301777.3287046}, $\hat \X_u^i$ and $\hat \X_v^i$ are fed into two fully-connected layers to obtain the dynamic weights $\w_u$ and $\w_v$, respectively. That is,
\begin{align}
    \label{eq:cat_he}
    \w_u = \softmax (\hat \W_2^{(u)} \relu(\hat \W_1^{(u)} \hat \X_u^i)) \\
    \w_v = \softmax (\hat \W_2^{(v)} \relu(\hat \W_1^{(v)} \hat \X_v^i) )
\end{align}
where $\hat \W_1^{(u)}$, $\hat \W_2^{(u)}$, $\hat \W_1^{(v)}$, and $\hat \W_2^{(v)}$ are the learnable parameters.

Different from \cite{He:2018:MCH:3301777.3287046}, we feed the feature vectors $\hat \X_u^i$, $\hat \X_v^i$ into two two independent PPNets to prevent overfitting, which can be formulated as follows:
\begin{align}
    \label{eq:ppnet_weight}
    \w_u = \softmax ({\rm{PPNe}}{{\rm{t}}_u} (\hat \X_u^i)) \\
    \w_v = \softmax ({\rm{PPNe}}{{\rm{t}}_v} (\hat \X_v^i))
\end{align}
Here, ${\rm{PPNe}}{{\rm{t}}_u}$, ${\rm{PPNe}}{{\rm{t}}_v}$ denote the two parameter prediction networks, which can be computed by Eq. (\ref{eq:ppnet}). We can get the final prediction as:
\begin{align}
    \label{eq:final}
    \hat y_o^i = {\w_u} \odot \hat y_u^i + {\w_v} \odot \hat y_v^i
\end{align}
where $\odot$ denotes the element-wise product operation.

Following \cite{He:2018:MCH:3301777.3287046}, the loss function is defined as follows:
\begin{align}
    \label {eq:total_loss}
    \L = L(\hat y_v^i, y^i) + L(\hat y_u^i, y^i) + L(\hat y_o^i, y^i)
\end{align}

\begin{table}
  \center
  \caption{Five datasets of Shanghai}
  \label{tab:datasets}
  \resizebox{0.45\textwidth}{!}{
  \begin{tabular}{lc}
    \toprule
    Dataset & \\
    \midrule
    Traffic accident data & \tabincell{c}{1,224,740 traffic accident records \\ from 2009 to 2016}  \\
    Taxi GPS trajectory data &  \tabincell{c}{50,000 taxies, each of which  \\ is sampled 1-2 times per minute}  \\
    POI Data & 1,363,709 POI records \\
    OpenStreetMap Data & \tabincell{c}{latitude from 30.7 degrees to 31.4 degrees, \\longitude from 121.1 degrees to 122.0 degrees} \\
    Satellite Imagery Data & 23,719 satellite images via Google Map API. \\
    \bottomrule
  \end{tabular}
  }
\end{table}

\subsection{Training}
We train our models by using the Adam solver \cite{kingma:adam} with a base learning rate of 1e-3, and dropout ratio of 0.4. We set the batch size and the training epochs to be 32 and 128, respectively. The cross-entropy loss function is adopted in our models. Furthermore, the early stopping technique is used to alleviate overfitting during training. Note that the training procedure will stop if the accuracy has not been improved in the last 16 epochs. We implement our models with PyTorch on Ubuntu 16.04.

\section{EXPERIMENTS}

\subsection{Datasets}
We evaluate our model with five datasets of Shanghai, which are present in Table \ref{tab:datasets}.

\begin{table}[t]
  \center
  \caption{Statistics of dimensions of different features}
  \label{tab:dims}
  \resizebox{0.42\textwidth}{!}{
  \begin{tabular}{ccccccc}
    \toprule
    Name & $D_{tra} $ &  $D_{poi}$  &   $D_{con}$ &  $D_{wid}$  & $D_{fra}$  & $D_{cnn}$       \\
    \midrule
    Dimension & 48 & 16 & 3 & 4 &   8 & 45  \\
    \bottomrule
  \end{tabular}
  }
\end{table}

\subsection{Methods Comparison}
To evaluate the performance of our model, we compare the following baseline methods including linear SVM, SVM with radial basis function kernel, Random Forests, gradient boosting decision tree (GBDT) \cite{Friedman00greedyfunction}, and fully-connected layer neural network (FCN).

\subsection{Experimental Settings}

In this experiment, we divide Shanghai (latitude from 30.7 degrees to 31.4 degrees, longitude from 121.1 degrees to 122.0 degrees) into 7,134 1km $\times$ 1km regions. The data point with outlying longitudes or latitudes is discarded. To alleviate the influence of class imbalance problem, resampling techniques are employed to preprocess the dataset. We randomly partition the dataset into the training set and the test set. The average results are reported based on \textbf{5-fold cross-validation}.

For the ResNet-152 model, satellite images are rescaled to 224 pixels $\times$ 224 pixels. For Feature-DFNN, we let $D_z = 64$, and we set the size $D_z$ to be 32 and 64 in ${\rm{PPNe}}{{\rm{t}}_u}$ and ${\rm{PPNe}}{{\rm{t}}_v}$, respectively. The statistics of the dimensions of different features are shown in Table \ref{tab:dims}.

\begin{table*}[t]
    \center
    \caption{Classification accuracies (\%) of all models under different features.}
    \label{tab:over_acc}
    \resizebox{0.63\textwidth}{!}{
    \begin{tabular}{cccc}
        \toprule
                     & Multimodal feature & Spatio-temporal feature & Visual feature \\
        \midrule
        Linear SVM          & 62.1   &  58.9 & 54.7    \\
        SVM with RBF kernel & 71.8   &  64.0 & 60.4     \\
        Random Forests      & 73.2   &  68.6 & 66.5    \\
        GDBT                & 74.5   &  69.8 & 65.9   \\
        FCN                 & 78.0   &  73.0 & 68.5   \\
        \midrule
        CNNE for classification \cite{He:2018:MCH:3301777.3287046} & 80.2 &  73.2  &  68.3 \\
        Feature-DFNN                & 81.6   &   74.5   & 69.1  \\
        Model-DFNN                & \textbf{83.0}   &   74.8   & 70.1 \\
        \bottomrule
    \end{tabular}
    }
\end{table*}

\subsection{Experimental Results and Analysis}

\subsubsection{Quantitative results}
\paragraph{Overall Evaluation}
We evaluate the performance of different models on the multimodal feature. 
The classification accuracies are illustrated in Table \ref{tab:over_acc}. First, Linear SVM shows the unsatisfactory performance, which is consistent with our expectation. A possible explanation for this might be that Linear SVM lacks the ability to fit into the non-linear multimodal data. GDBT and Random Forests perform better than the SVM models with linear and RBF kernels as they have powerful non-linear data fitting capabilities, especially for the multimodal data. FCN is a strong baseline model, which outperforms GDBT with 3.5\% performance improvement as FCN benefits from end-to-end data-driven learning. Feature-DFNN outperforms CNNE as Feature-DFNN gains improvement by fusing the features as mutual interactions, and the matrix decomposition technique alleviates overfitting. Model-DFNN performs better than Feature-DFNN, achieving the best performance, which demonstrates the power of fusing cross-domain data at model level. In addition, incorporating the matrix decomposition technique makes Model-DFNN performs better than CNNE due to the advance of reducing the number of parameters to a more compact representation to alleviate overfitting. 


\begin{table}
    \center
    \caption{Accuracies (\%) of Feature-DFNN based on different features}
    \label{tab:feature_dfnn}
    \resizebox{0.42\textwidth}{!}{
    \begin{tabular}{ccc}
      \toprule
                &  Input features   &   Accuracy          \\
      \midrule
      Classification network &  Spatio-temporal feature &   \multirow{2}*{81.6}   \\
      PPNet  &  Visual feature &                                      \\
      \midrule
      Classification network & Visual feature & \multirow{2}*{78.8}    \\
      PPNet &   Spatio-temporal feature   &     \\
      \bottomrule
    \end{tabular}
    }
\end{table}

\paragraph{Evaluation on Single-Modality Data}
%
%

To evaluate the benefit of applying multimodal information fusion, we report the performance of using the feature of either modality alone for the sake of comparison. The results are shown in Table \ref{tab:over_acc}. We reach the following findings:
\begin{itemize}
  \item Notably, the models achieve better accuracies based on the multimodal feature than those based on single-modality data (the spatio-temporal feature or visual feature). This demonstrates that the fused features can provide complementary information to improve the overall performance in predicting traffic accident risk.
      Besides, the performance improvement promised by the proposed model confirms its power in multimodal information fusion.

  \item It is interesting to note that the models based on spatio-temporal feature perform better than the ones based on visual feature. These results can be attributed to the high-level semantic spatio-temporal feature, which should be more explicitly correlated to the traffic accident risk.
\end{itemize}

\paragraph{Evaluation on Feature-DFNN}
Feature-DFNN takes the data of a single-modality as input and generates dynamic parameters for the data of another modality to achieve multimodal data fusion. Here, we conduct an experiment to see whether feeding different features into PPNet (classification network) can affect the prediction accuracy. The accuracies are shown in Table \ref{tab:feature_dfnn}. From Table \ref{tab:feature_dfnn}, we can see that one case (classification network based on the spatio-temporal feature and PPNet based on the visual feature) achieves better performance than the other (classification network based on the visual feature and PPNet based on the spatio-temporal feature). A possible explanation for this result is that spatio-temporal feature acts as the primary decision basis due to its explicitly physical meaning in association with the risk of traffic accidents.


\subsubsection{Qualitative analysis}
\label{sec:vis}
\begin{figure}[t]
  \centering
  \subfigure[Ground truth.]{
    \label{fig:risk1}
    \includegraphics[width=0.21\textwidth, height=0.21\textwidth]{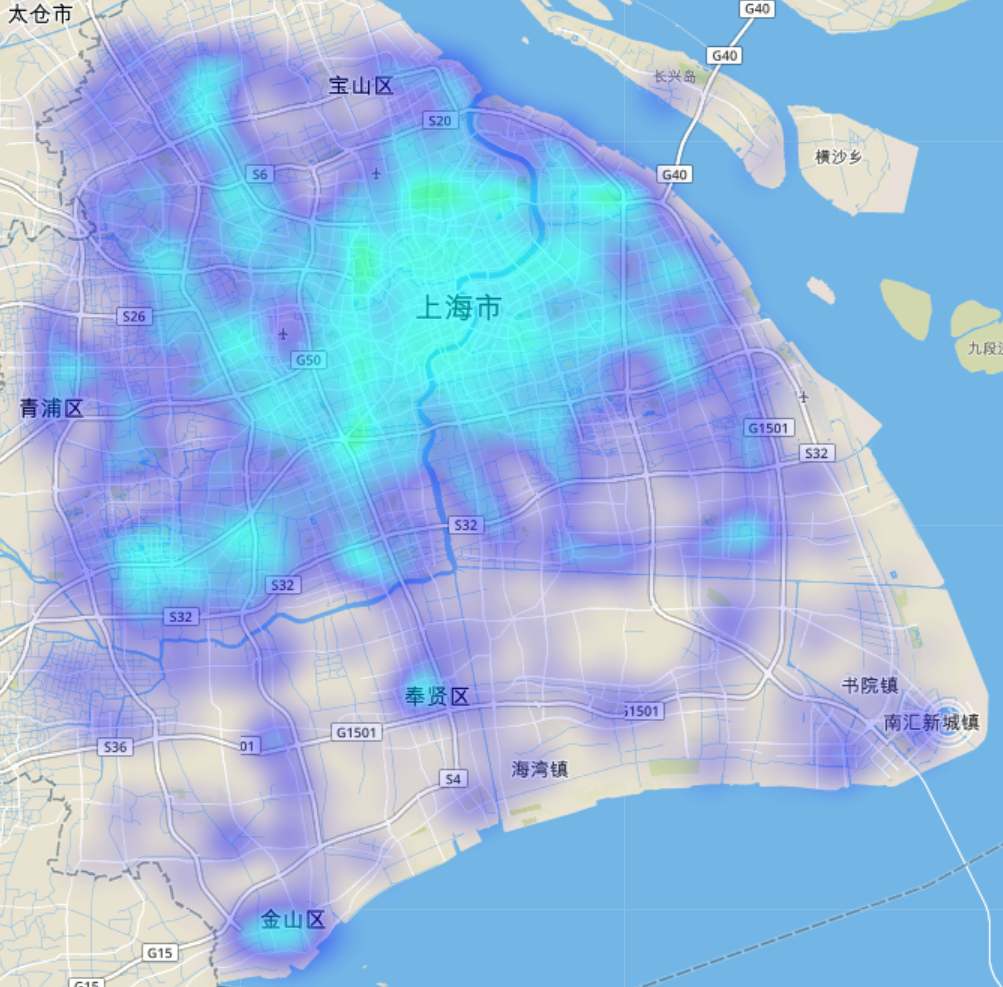}
  }
  \subfigure[Prediction.]{
    \label{fig:risk2}
    \includegraphics[width=0.21\textwidth, height=0.21\textwidth]{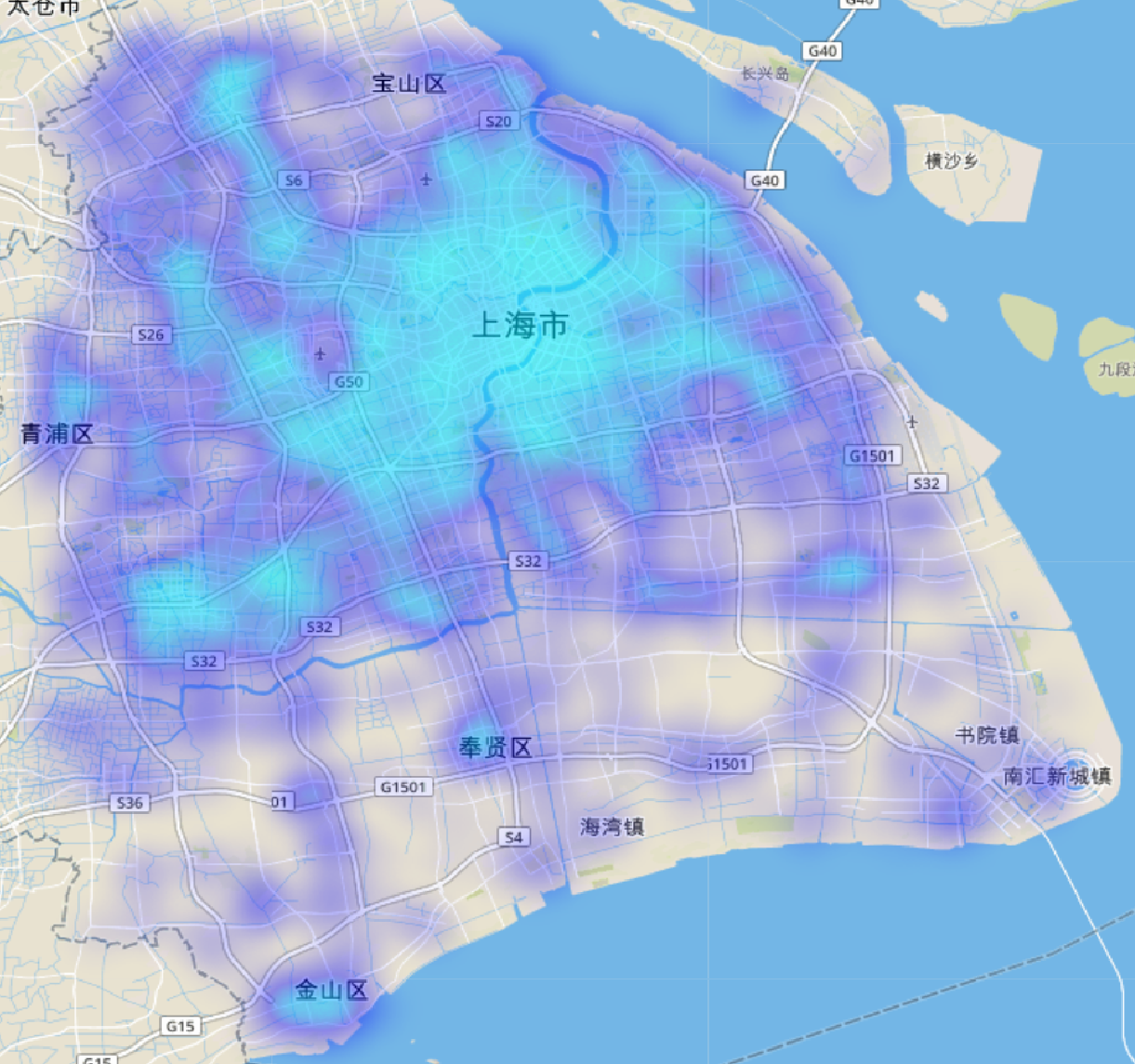}
  }
  \caption{Heatmaps of the traffic accident risk in Shanghai marked by color from light to dark to indicate the risk in descending order. (a) Ground truth. (b) Prediction.}
  \label{fg:city_risk}
\end{figure}

\begin{figure*}
    \centering
    \subfigure[]{
    \label{fig:high_1}
    \begin{minipage}[b]{0.14\linewidth}
    \includegraphics[width=1\linewidth]{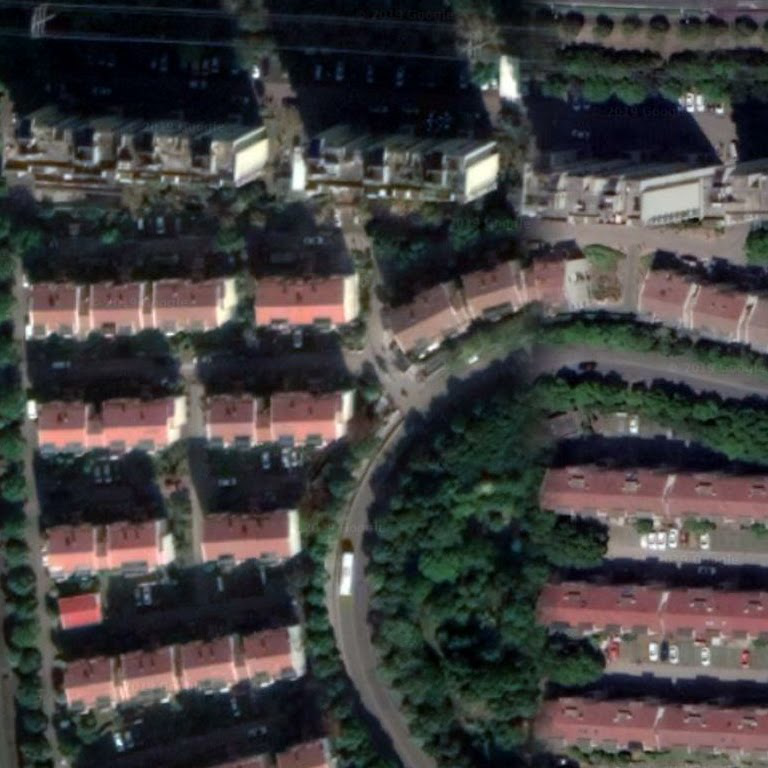}\vspace{1pt}
    \includegraphics[width=1\linewidth]{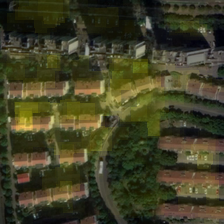}
    \end{minipage}
    }
    \subfigure[]{
    \label{fig:high_2}
    \begin{minipage}[b]{0.14\linewidth}
    \includegraphics[width=1\linewidth]{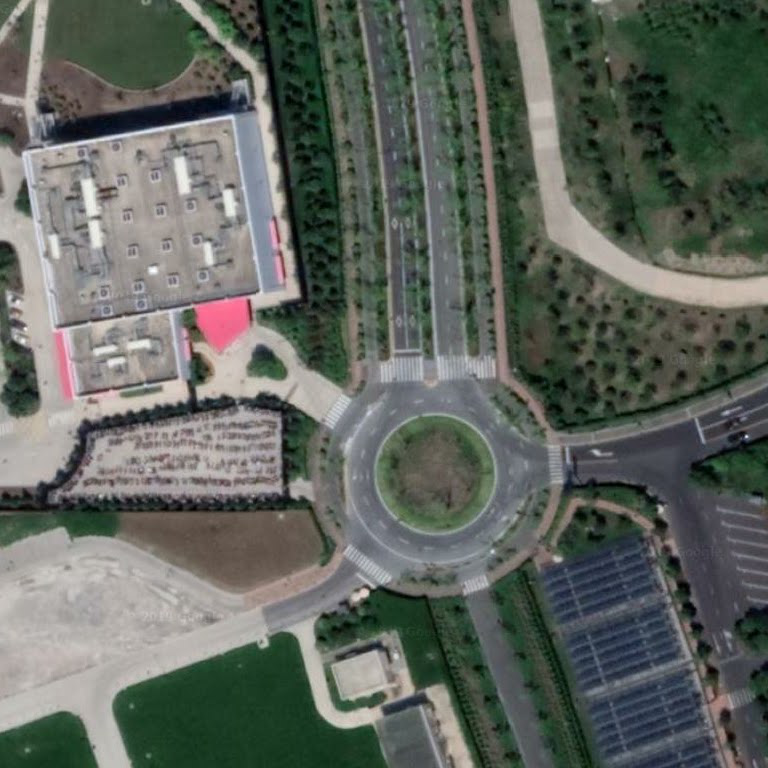}\vspace{1pt}
    \includegraphics[width=1\linewidth]{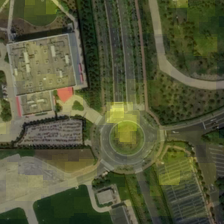}
    \end{minipage}
    }
    \subfigure[]{
    \label{fig:high_3}
    \begin{minipage}[b]{0.14\linewidth}
    \includegraphics[width=1\linewidth]{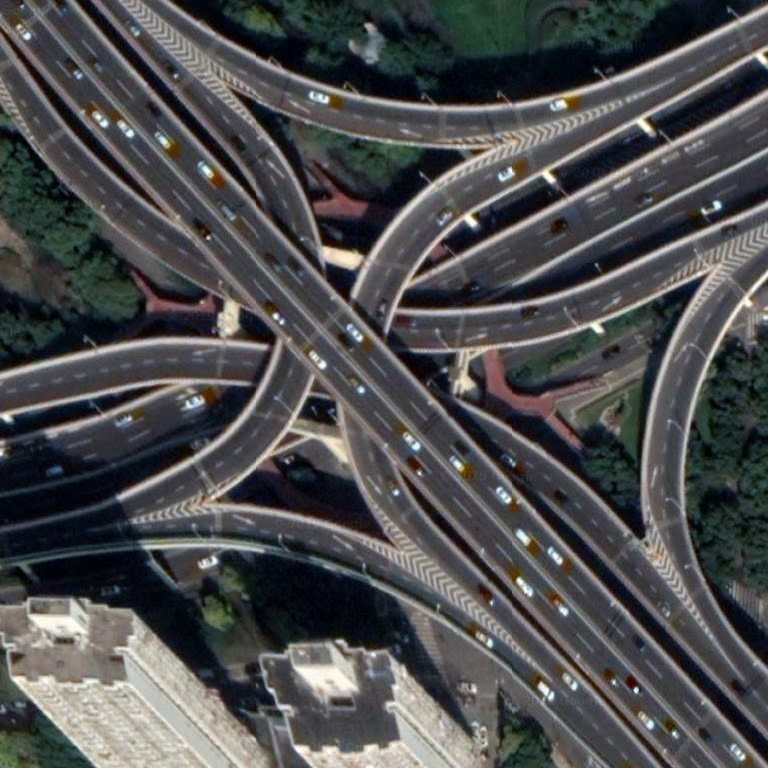}\vspace{1pt}
    \includegraphics[width=1\linewidth]{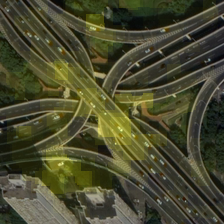}
    \end{minipage}
    }
    \subfigure[]{
    \label{fig:high_4}
    \begin{minipage}[b]{0.14\linewidth}
    \includegraphics[width=1\linewidth]{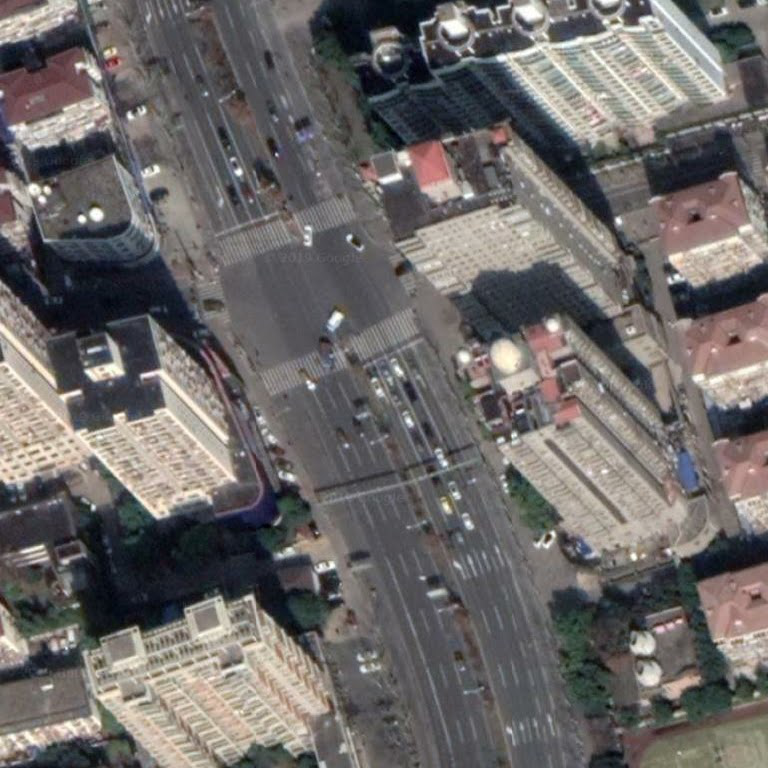}\vspace{1pt}
    \includegraphics[width=1\linewidth]{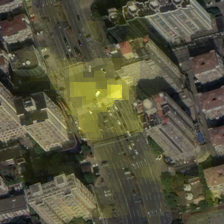}
    \end{minipage}
    }
    \subfigure[]{
    \label{fig:high_5}
    \begin{minipage}[b]{0.14\linewidth}
    \includegraphics[width=1\linewidth]{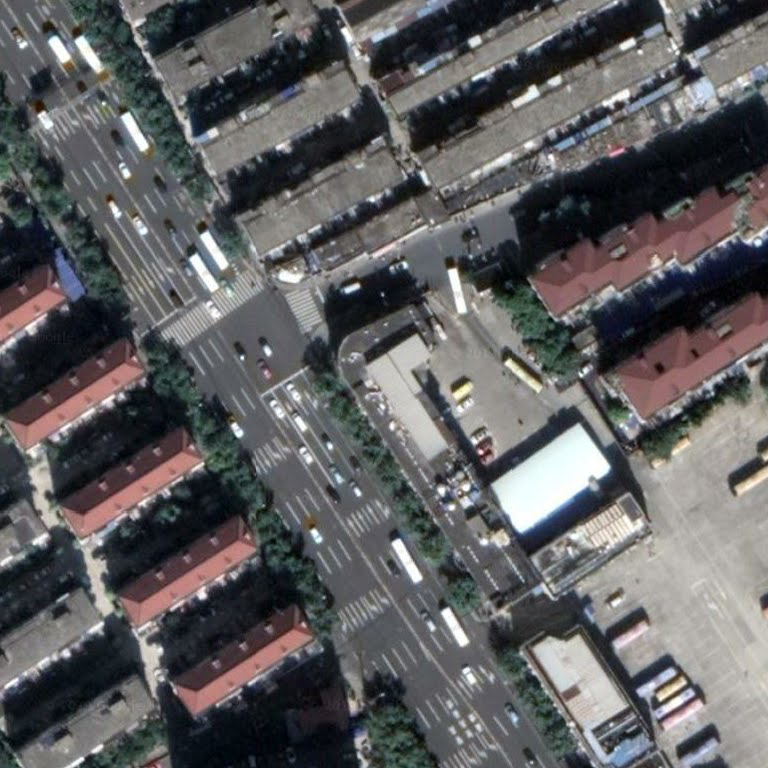}\vspace{1pt}
    \includegraphics[width=1\linewidth]{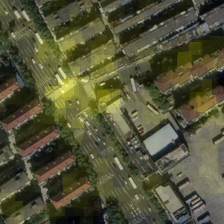}
    \end{minipage}
    }
    \subfigure[]{
    \label{fig:high_6}
    \begin{minipage}[b]{0.14\linewidth}
    \includegraphics[width=1\linewidth]{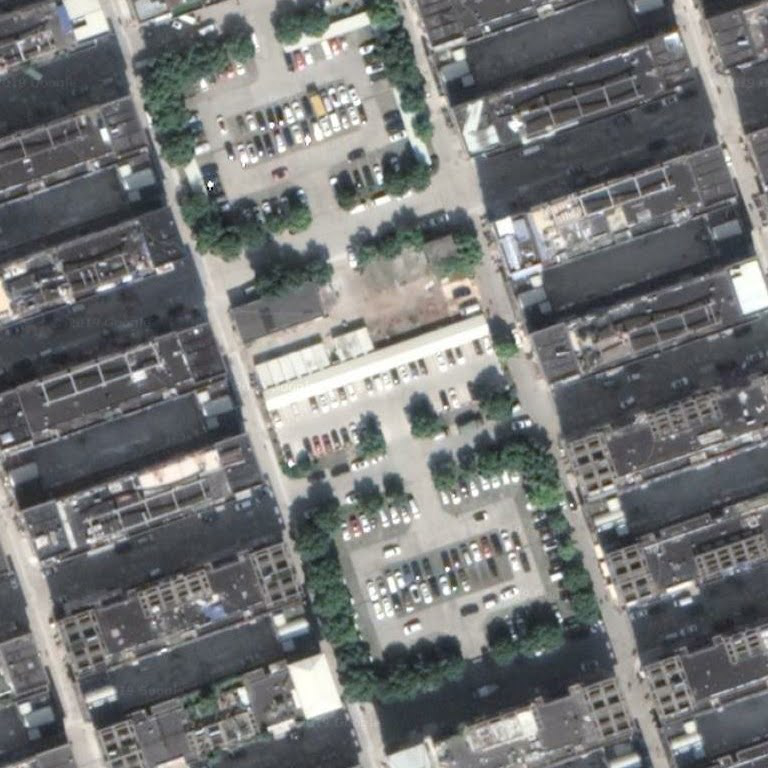}\vspace{1pt}
    \includegraphics[width=1\linewidth]{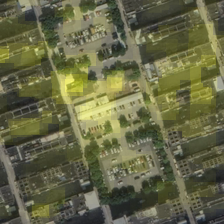}
    \end{minipage}
    }
    \caption{Visual cues related to the high traffic accident risk, where the pixels corresponding with high risk are highlighted (yellow) in the second row and the first row contains the original image patches.}
    \label{fg:vis_high}
\end{figure*}

In this section, we perform visual analysis from the perspective of macro and micro to allow straightforward insight into the results of risk prediction on traffic accidents. First, we visualize the traffic accident risk at the city-scale in the form of a heatmap, which is shown in Figure \ref{fg:city_risk}. In Figure \ref{fg:city_risk}, we found that the predicted heatmap (Figure \ref{fig:risk2}) resulting from the best Model-DFNN is similar to that of the ground truth (Figure \ref{fig:risk1}). Such a traffic accident risk heatmap provides a valuable reference for emergency response when planning resource allocation. It is known that planning the locations to allocate resources for possible emergency response to abnormal events is a optimization problem. However, city-scale risk map is the basis for decision making in terms of doing such planning. Intuitively, the proposed model can be applied to newly developed regions to predict accident-prone locations, where the historical data are rare or do not exist at all.

In addition, we present an elaborate qualitative analysis of our models to get an intuitive understanding of the causes of traffic accidents. In other words, we aim to find out what visual cues contribute to high or low traffic accident risk. We make use of the integrated gradients \cite{Sundararajan:2017:AAD:3305890.3306024} technique, which is simple to implement and requires no modification to the original neural model. Specifically, it is employed to investigate pixel importance in the prediction made by Model-DFNN. First, we compute the gradients for the output of the highest-scoring class concerning the pixels of the input satellite image. Then, we overlay integrated gradients on the actual satellite image to highlight the image regions.

In order to find out the visual patterns that correlated to the high traffic accident risk, we randomly select the test examples for which Model-DFNN correctly predicts, as illustrated in Figure \ref{fg:vis_high}. From Figure \ref{fg:vis_high}, we can see that six original images and the corresponding images with highlighted pixels to be reported as high-risk class members by Model-DFNN. We find that the places where traffic accidents are likely to occur have in general complex structures according to Figure \ref{fg:vis_high}(a), (b), and (c), which are consistent with our expectation. Another place where the risk of a traffic accident is high is the crossroads in the business district (Figure \ref{fig:high_4}) and the crossroads in the residential areas (Figure \ref{fig:high_5}). These places are characterized by a large number of vehicles and people. Furthermore, the narrow road in the residential area is easy to cause traffic accident as shown in Figure \ref{fig:high_6}. The above observations indicate that the tips for slowing down should be set to alarm drivers in high-risk regions from the perspective of traffic management. Furthermore, we should avoid these road network designs that are prone to traffic accidents during urban planning.

\begin{figure*}[t]
    \centering
    \subfigure[]{
    \label{fig:low_1}
    \begin{minipage}[b]{0.14\linewidth}
    \includegraphics[width=1\linewidth]{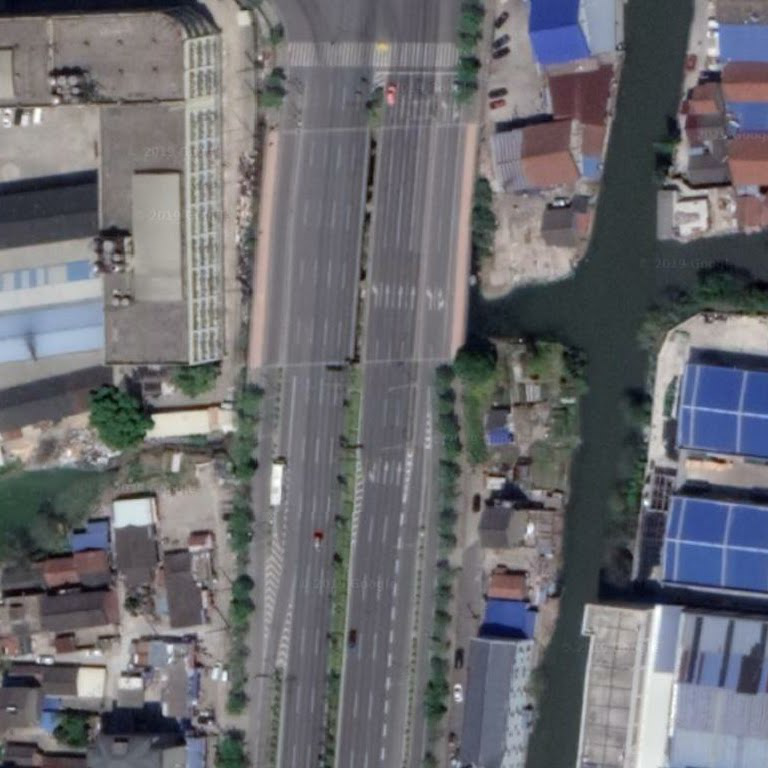}\vspace{1pt}
    \includegraphics[width=1\linewidth]{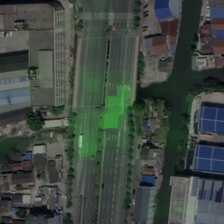}
    \end{minipage}
    }
    \subfigure[]{
    \label{fig:low_2}
    \begin{minipage}[b]{0.14\linewidth}
    \includegraphics[width=1\linewidth]{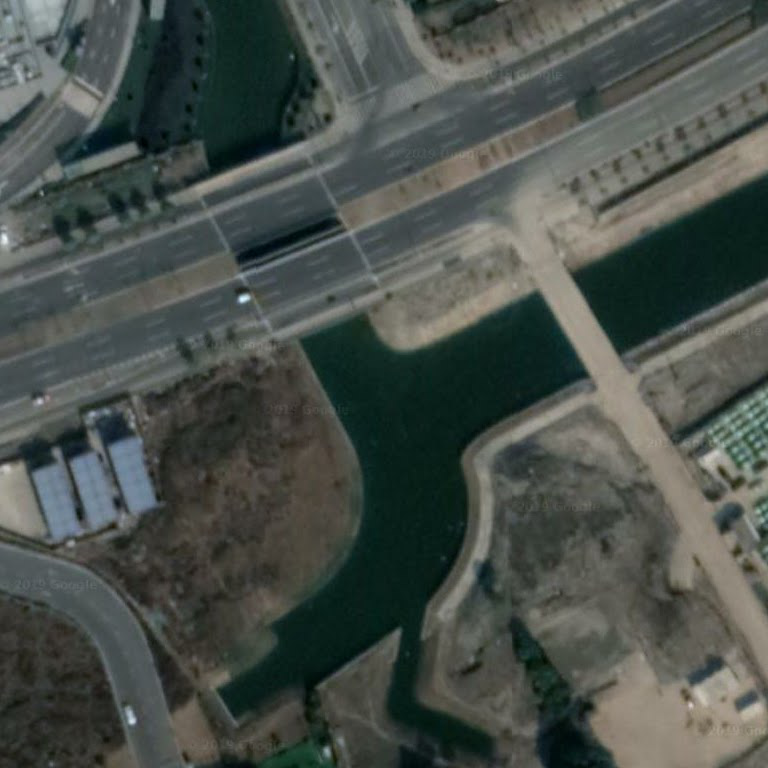}\vspace{1pt}
    \includegraphics[width=1\linewidth]{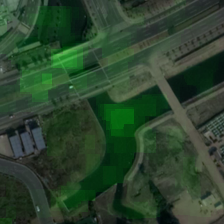}
    \end{minipage}
    }
    \subfigure[]{
    \label{fig:low_3}
    \begin{minipage}[b]{0.14\linewidth}
    \includegraphics[width=1\linewidth]{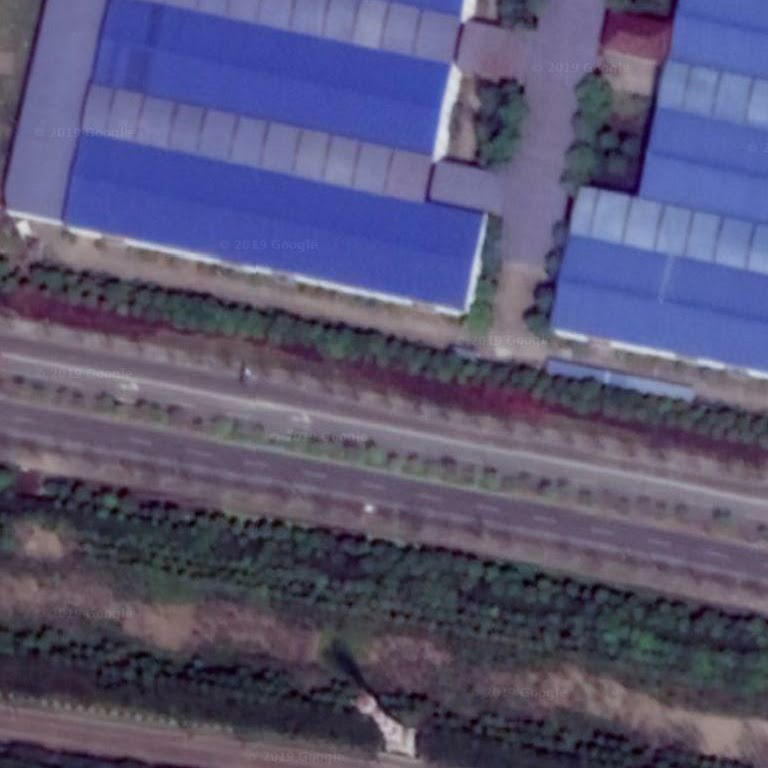}\vspace{1pt}
    \includegraphics[width=1\linewidth]{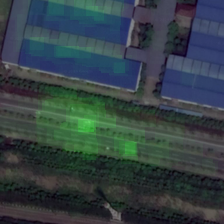}
    \end{minipage}
    }
    \subfigure[]{
    \label{fig:low_4}
    \begin{minipage}[b]{0.14\linewidth}
    \includegraphics[width=1\linewidth]{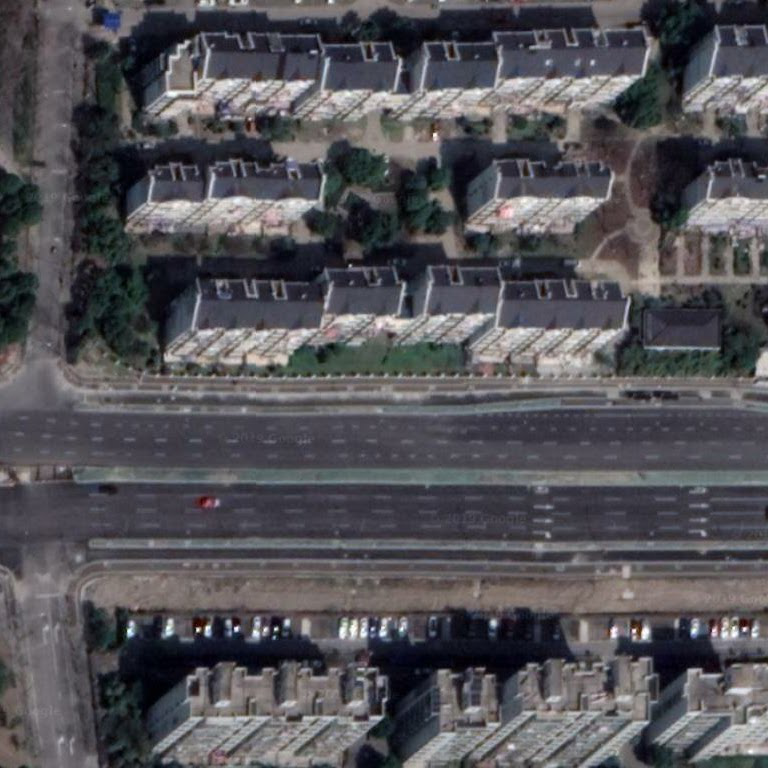}\vspace{1pt}
    \includegraphics[width=1\linewidth]{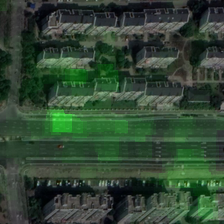}
    \end{minipage}
    }
    \subfigure[]{
    \label{fig:low_5}
    \begin{minipage}[b]{0.14\linewidth}
    \includegraphics[width=1\linewidth]{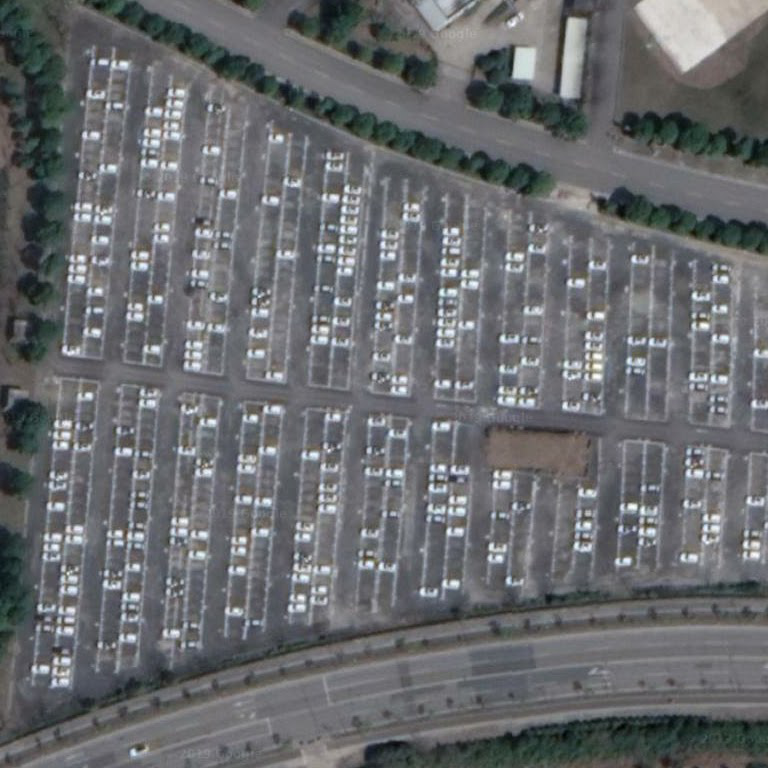}\vspace{1pt}
    \includegraphics[width=1\linewidth]{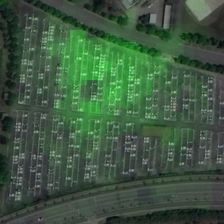}
    \end{minipage}
    }
    \subfigure[]{
    \label{fig:low_6}
    \begin{minipage}[b]{0.14\linewidth}
    \includegraphics[width=1\linewidth]{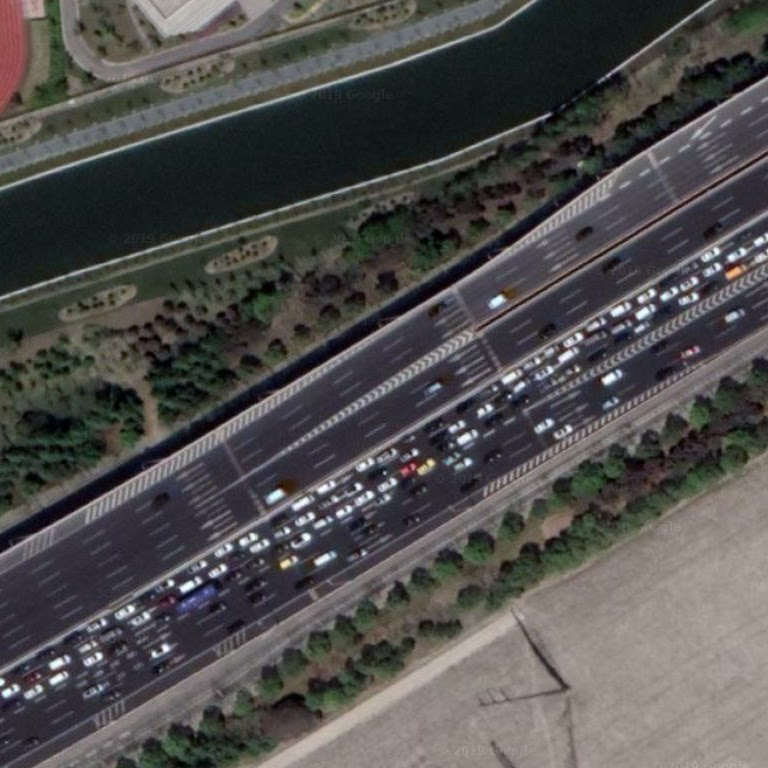}\vspace{1pt}
    \includegraphics[width=1\linewidth]{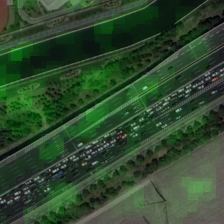}
    \end{minipage}
    }
    \caption{Visual cues related to the low traffic accident risk, where the pixels corresponding with low risk are highlighted (green) in the second row and the first row contains the original image patches.}
    \label{fg:vis_low}
\end{figure*}

Besides, we also investigate the visual cues that correlated to the low traffic accident risk. Six examples are selected from the testing set, which are shown in Figure \ref{fg:vis_low}. First, we found that the straightforward and spacious roads are not prone to traffic accidents (Figure \ref{fig:low_1}-\ref{fig:low_4}), which is consistent with our common sense. In Figure \ref{fig:low_5}, the parking area is identified by Model-DFNN as low risk. A possible explanation for this might be that a well-managed and speed-limited parking lot is not prone to traffic accidents. Not surprisingly, the road without cars receives less risk score than the one with cars in Figure \ref{fig:low_6}. 

In summary, these visual patterns can visually explain the causes of traffic accidents and help traffic management, city planning, and emergency response.

\section{CONCLUSION}

In this paper, we investigate traffic accident risk prediction from a new point of view by fusing multimodal features in decision making. First, we extract the spatio-temporal feature and the visual feature from multiple data sources: Taxi GPS data, POI data, OpenStreetMap, and satellite imagery data. Afterwards, we realize two multimodal information fusion neural models referred to as Feature-DFNN and Model-DFNN with dynamic weighting adaptive to the context on the fly, and develop the models into more robust forms by using matrix decomposition to reduce the original model to a more compact representation against overfitting. We conduct extensive experiments and the results demonstrate the power of the models in fusing multimodal data. Furthermore, we employ the integrated gradients method to interpret the results and find out the causes of traffic accidents from both macro and micro points of view, which is helpful for city-planning, traffic management, and emergency response.

\ifCLASSOPTIONcaptionsoff
  \newpage
\fi



\bibliographystyle{IEEEtran}
\bibliography{ecai,ecai2}

\begin{thebibliography}{10}
\providecommand{\url}[1]{#1}
\csname url@samestyle\endcsname
\providecommand{\newblock}{\relax}
\providecommand{\bibinfo}[2]{#2}
\providecommand{\BIBentrySTDinterwordspacing}{\spaceskip=0pt\relax}
\providecommand{\BIBentryALTinterwordstretchfactor}{4}
\providecommand{\BIBentryALTinterwordspacing}{\spaceskip=\fontdimen2\font plus
\BIBentryALTinterwordstretchfactor\fontdimen3\font minus
  \fontdimen4\font\relax}
\providecommand{\BIBforeignlanguage}[2]{{%
\expandafter\ifx\csname l@#1\endcsname\relax
\typeout{** WARNING: IEEEtran.bst: No hyphenation pattern has been}%
\typeout{** loaded for the language `#1'. Using the pattern for}%
\typeout{** the default language instead.}%
\else
\language=\csname l@#1\endcsname
\fi
#2}}
\providecommand{\BIBdecl}{\relax}
\BIBdecl

\bibitem{Zheng:2014:UCC:2648782.2629592}
\BIBentryALTinterwordspacing
Y.~Zheng, L.~Capra, O.~Wolfson, and H.~Yang, ``Urban computing: Concepts,
  methodologies, and applications,'' \emph{ACM Trans. Intell. Syst. Technol.},
  vol.~5, no.~3, pp. 38:1--38:55, Sep. 2014. [Online]. Available:
  \url{http://doi.acm.org/10.1145/2629592}
\BIBentrySTDinterwordspacing

\bibitem{Chen:2018:RRO:3178157.3161159}
\BIBentryALTinterwordspacing
L.~Chen, X.~Fan, L.~Wang, D.~Zhang, Z.~Yu, J.~Li, T.-M.-T. Nguyen, G.~Pan, and
  C.~Wang, ``Radar: Road obstacle identification for disaster response
  leveraging cross-domain urban data,'' \emph{Proc. ACM Interact. Mob. Wearable
  Ubiquitous Technol.}, vol.~1, no.~4, pp. 130:1--130:23, Jan. 2018. [Online].
  Available: \url{http://doi.acm.org/10.1145/3161159}
\BIBentrySTDinterwordspacing

\bibitem{Karamshuk:2013:GMO:2487575.2487616}
\BIBentryALTinterwordspacing
D.~Karamshuk, A.~Noulas, S.~Scellato, V.~Nicosia, and C.~Mascolo,
  ``Geo-spotting: Mining online location-based services for optimal retail
  store placement,'' in \emph{Proceedings of the 19th ACM SIGKDD International
  Conference on Knowledge Discovery and Data Mining}, ser. KDD '13.\hskip 1em
  plus 0.5em minus 0.4em\relax New York, NY, USA: ACM, 2013, pp. 793--801.
  [Online]. Available: \url{http://doi.acm.org/10.1145/2487575.2487616}
\BIBentrySTDinterwordspacing

\bibitem{Naik7571}
\BIBentryALTinterwordspacing
N.~Naik, S.~D. Kominers, R.~Raskar, E.~L. Glaeser, and C.~A. Hidalgo,
  ``Computer vision uncovers predictors of physical urban change,''
  \emph{Proceedings of the National Academy of Sciences}, vol. 114, no.~29, pp.
  7571--7576, 2017. [Online]. Available:
  \url{https://www.pnas.org/content/114/29/7571}
\BIBentrySTDinterwordspacing

\bibitem{Gebru13108}
\BIBentryALTinterwordspacing
T.~Gebru, J.~Krause, Y.~Wang, D.~Chen, J.~Deng, E.~L. Aiden, and L.~Fei-Fei,
  ``Using deep learning and google street view to estimate the demographic
  makeup of neighborhoods across the united states,'' \emph{Proceedings of the
  National Academy of Sciences}, vol. 114, no.~50, pp. 13\,108--13\,113, 2017.
  [Online]. Available: \url{https://www.pnas.org/content/114/50/13108}
\BIBentrySTDinterwordspacing

\bibitem{Porzi:2015:PUU:2733373.2806273}
\BIBentryALTinterwordspacing
L.~Porzi, S.~Rota~Bul\`{o}, B.~Lepri, and E.~Ricci, ``Predicting and
  understanding urban perception with convolutional neural networks,'' in
  \emph{Proceedings of the 23rd ACM International Conference on Multimedia},
  ser. MM '15.\hskip 1em plus 0.5em minus 0.4em\relax New York, NY, USA: ACM,
  2015, pp. 139--148. [Online]. Available:
  \url{http://doi.acm.org/10.1145/2733373.2806273}
\BIBentrySTDinterwordspacing

\bibitem{6875954}
S.~M. {Arietta}, A.~A. {Efros}, R.~{Ramamoorthi}, and M.~{Agrawala}, ``City
  forensics: Using visual elements to predict non-visual city attributes,''
  \emph{IEEE Transactions on Visualization and Computer Graphics}, vol.~20,
  no.~12, pp. 2624--2633, Dec 2014.

\bibitem{Quercia:2014:ACM:2531602.2531613}
\BIBentryALTinterwordspacing
D.~Quercia, N.~K. O'Hare, and H.~Cramer, ``Aesthetic capital: What makes london
  look beautiful, quiet, and happy?'' in \emph{Proceedings of the 17th ACM
  Conference on Computer Supported Cooperative Work \&\#38; Social Computing},
  ser. CSCW '14.\hskip 1em plus 0.5em minus 0.4em\relax New York, NY, USA: ACM,
  2014, pp. 945--955. [Online]. Available:
  \url{http://doi.acm.org/10.1145/2531602.2531613}
\BIBentrySTDinterwordspacing

\bibitem{10.1007/978-3-319-46448-0_12}
A.~Dubey, N.~Naik, D.~Parikh, R.~Raskar, and C.~A. Hidalgo, ``Deep learning the
  city: Quantifying urban perception at a global scale,'' in \emph{Computer
  Vision -- ECCV 2016}, B.~Leibe, J.~Matas, N.~Sebe, and M.~Welling, Eds.\hskip
  1em plus 0.5em minus 0.4em\relax Cham: Springer International Publishing,
  2016, pp. 196--212.

\bibitem{DeNadai:2016:SLN:2964284.2964312}
\BIBentryALTinterwordspacing
M.~De~Nadai, R.~L. Vieriu, G.~Zen, S.~Dragicevic, N.~Naik, M.~Caraviello, C.~A.
  Hidalgo, N.~Sebe, and B.~Lepri, ``Are safer looking neighborhoods more
  lively?: A multimodal investigation into urban life,'' in \emph{Proceedings
  of the 24th ACM International Conference on Multimedia}, ser. MM '16.\hskip
  1em plus 0.5em minus 0.4em\relax New York, NY, USA: ACM, 2016, pp.
  1127--1135. [Online]. Available:
  \url{http://doi.acm.org/10.1145/2964284.2964312}
\BIBentrySTDinterwordspacing

\bibitem{Jean790}
\BIBentryALTinterwordspacing
N.~Jean, M.~Burke, M.~Xie, W.~M. Davis, D.~B. Lobell, and S.~Ermon, ``Combining
  satellite imagery and machine learning to predict poverty,'' \emph{Science},
  vol. 353, no. 6301, pp. 790--794, 2016. [Online]. Available:
  \url{https://science.sciencemag.org/content/353/6301/790}
\BIBentrySTDinterwordspacing

\bibitem{He:2018:MCH:3301777.3287046}
\BIBentryALTinterwordspacing
Z.~He and S.~Yang, ``Multi-view commercial hotness prediction using
  context-aware neural network ensemble,'' \emph{Proc. ACM Interact. Mob.
  Wearable Ubiquitous Technol.}, vol.~2, no.~4, pp. 168:1--168:19, Dec. 2018.
  [Online]. Available: \url{http://doi.acm.org/10.1145/3287046}
\BIBentrySTDinterwordspacing

\bibitem{CHANG2005541}
\BIBentryALTinterwordspacing
L.-Y. Chang, ``Analysis of freeway accident frequencies: Negative binomial
  regression versus artificial neural network,'' \emph{Safety Science},
  vol.~43, no.~8, pp. 541 -- 557, 2005. [Online]. Available:
  \url{http://www.sciencedirect.com/science/article/pii/S0925753505000676}
\BIBentrySTDinterwordspacing

\bibitem{LIN2015444}
\BIBentryALTinterwordspacing
L.~Lin, Q.~Wang, and A.~W. Sadek, ``A novel variable selection method based on
  frequent pattern tree for real-time traffic accident risk prediction,''
  \emph{Transportation Research Part C: Emerging Technologies}, vol.~55, pp.
  444 -- 459, 2015, engineering and Applied Sciences Optimization (OPT-i) -
  Professor Matthew G. Karlaftis Memorial Issue. [Online]. Available:
  \url{http://www.sciencedirect.com/science/article/pii/S0968090X15000947}
\BIBentrySTDinterwordspacing

\bibitem{CHANG2005365}
\BIBentryALTinterwordspacing
L.-Y. Chang and W.-C. Chen, ``Data mining of tree-based models to analyze
  freeway accident frequency,'' \emph{Journal of Safety Research}, vol.~36,
  no.~4, pp. 365 -- 375, 2005. [Online]. Available:
  \url{http://www.sciencedirect.com/science/article/pii/S0022437505000708}
\BIBentrySTDinterwordspacing

\bibitem{CALIENDO2007657}
\BIBentryALTinterwordspacing
C.~Caliendo, M.~Guida, and A.~Parisi, ``A crash-prediction model for multilane
  roads,'' \emph{Accident Analysis \& Prevention}, vol.~39, no.~4, pp. 657 --
  670, 2007. [Online]. Available:
  \url{http://www.sciencedirect.com/science/article/pii/S0001457506001965}
\BIBentrySTDinterwordspacing

\bibitem{EISENBERG2004637}
\BIBentryALTinterwordspacing
D.~Eisenberg, ``The mixed effects of precipitation on traffic crashes,''
  \emph{Accident Analysis \& Prevention}, vol.~36, no.~4, pp. 637 -- 647, 2004.
  [Online]. Available:
  \url{http://www.sciencedirect.com/science/article/pii/S000145750300085X}
\BIBentrySTDinterwordspacing

\bibitem{BERGELHAYAT2013456}
\BIBentryALTinterwordspacing
R.~Bergel-Hayat, M.~Debbarh, C.~Antoniou, and G.~Yannis, ``Explaining the road
  accident risk: Weather effects,'' \emph{Accident Analysis \& Prevention},
  vol.~60, pp. 456 -- 465, 2013. [Online]. Available:
  \url{http://www.sciencedirect.com/science/article/pii/S0001457513000948}
\BIBentrySTDinterwordspacing

\bibitem{Chen:2016:LDR:3015812.3015863}
\BIBentryALTinterwordspacing
Q.~Chen, X.~Song, H.~Yamada, and R.~Shibasaki, ``Learning deep representation
  from big and heterogeneous data for traffic accident inference,'' in
  \emph{Proceedings of the Thirtieth AAAI Conference on Artificial
  Intelligence}, ser. AAAI'16.\hskip 1em plus 0.5em minus 0.4em\relax AAAI
  Press, 2016, pp. 338--344. [Online]. Available:
  \url{http://dl.acm.org/citation.cfm?id=3015812.3015863}
\BIBentrySTDinterwordspacing

\bibitem{Yuan:2018:HDL:3219819.3219922}
\BIBentryALTinterwordspacing
Z.~Yuan, X.~Zhou, and T.~Yang, ``Hetero-convlstm: A deep learning approach to
  traffic accident prediction on heterogeneous spatio-temporal data,'' in
  \emph{Proceedings of the 24th ACM SIGKDD International Conference on
  Knowledge Discovery \&\#38; Data Mining}, ser. KDD '18.\hskip 1em plus 0.5em
  minus 0.4em\relax New York, NY, USA: ACM, 2018, pp. 984--992. [Online].
  Available: \url{http://doi.acm.org/10.1145/3219819.3219922}
\BIBentrySTDinterwordspacing

\bibitem{Najjar:2017:CSI:3298023.3298224}
\BIBentryALTinterwordspacing
A.~Najjar, S.~Kaneko, and Y.~Miyanaga, ``Combining satellite imagery and open
  data to map road safety,'' in \emph{Proceedings of the Thirty-First AAAI
  Conference on Artificial Intelligence}, ser. AAAI'17.\hskip 1em plus 0.5em
  minus 0.4em\relax AAAI Press, 2017, pp. 4524--4530. [Online]. Available:
  \url{http://dl.acm.org/citation.cfm?id=3298023.3298224}
\BIBentrySTDinterwordspacing

\bibitem{8397033}
Q.~{Chen}, X.~{Song}, Z.~{Fan}, T.~{Xia}, H.~{Yamada}, and R.~{Shibasaki}, ``A
  context-aware nonnegative matrix factorization framework for traffic accident
  risk estimation via heterogeneous data,'' in \emph{2018 IEEE Conference on
  Multimedia Information Processing and Retrieval (MIPR)}, April 2018, pp.
  346--351.

\bibitem{Yang:2017:PCA:3139486.3130983}
\BIBentryALTinterwordspacing
S.~Yang, M.~Wang, W.~Wang, Y.~Sun, J.~Gao, W.~Zhang, and J.~Zhang, ``Predicting
  commercial activeness over urban big data,'' \emph{Proc. ACM Interact. Mob.
  Wearable Ubiquitous Technol.}, vol.~1, no.~3, pp. 119:1--119:20, Sep. 2017.
  [Online]. Available: \url{http://doi.acm.org/10.1145/3130983}
\BIBentrySTDinterwordspacing

\bibitem{Sundararajan:2017:AAD:3305890.3306024}
\BIBentryALTinterwordspacing
M.~Sundararajan, A.~Taly, and Q.~Yan, ``Axiomatic attribution for deep
  networks,'' in \emph{Proceedings of the 34th International Conference on
  Machine Learning - Volume 70}, ser. ICML'17.\hskip 1em plus 0.5em minus
  0.4em\relax JMLR.org, 2017, pp. 3319--3328. [Online]. Available:
  \url{http://dl.acm.org/citation.cfm?id=3305890.3306024}
\BIBentrySTDinterwordspacing

\bibitem{ABELLAN20136047}
\BIBentryALTinterwordspacing
J.~Abellán, G.~López, and J.~de~Oña, ``Analysis of traffic accident severity
  using decision rules via decision trees,'' \emph{Expert Systems with
  Applications}, vol.~40, no.~15, pp. 6047 -- 6054, 2013. [Online]. Available:
  \url{http://www.sciencedirect.com/science/article/pii/S0957417413003138}
\BIBentrySTDinterwordspacing

\bibitem{DBLP:journals/corr/abs-1710-09543}
\BIBentryALTinterwordspacing
H.~Ren, Y.~Song, J.~Liu, Y.~Hu, and J.~Lei, ``A deep learning approach to the
  prediction of short-term traffic accident risk,'' \emph{CoRR}, vol.
  abs/1710.09543, 2017. [Online]. Available:
  \url{http://arxiv.org/abs/1710.09543}
\BIBentrySTDinterwordspacing

\bibitem{Yuan:2012:DRD:2339530.2339561}
\BIBentryALTinterwordspacing
J.~Yuan, Y.~Zheng, and X.~Xie, ``Discovering regions of different functions in
  a city using human mobility and pois,'' in \emph{Proceedings of the 18th ACM
  SIGKDD International Conference on Knowledge Discovery and Data Mining}, ser.
  KDD '12.\hskip 1em plus 0.5em minus 0.4em\relax New York, NY, USA: ACM, 2012,
  pp. 186--194. [Online]. Available:
  \url{http://doi.acm.org/10.1145/2339530.2339561}
\BIBentrySTDinterwordspacing

\bibitem{4767591}
A.~P. {Pentland}, ``Fractal-based description of natural scenes,'' \emph{IEEE
  Transactions on Pattern Analysis and Machine Intelligence}, vol. PAMI-6,
  no.~6, pp. 661--674, Nov 1984.

\bibitem{jean2016combining}
N.~Jean, M.~Burke, M.~Xie, W.~M. Davis, D.~B. Lobell, and S.~Ermon, ``Combining
  satellite imagery and machine learning to predict poverty,'' \emph{Science},
  vol. 353, no. 6301, pp. 790--794, 2016.

\bibitem{Wang:2018:UPC:3184558.3186581}
\BIBentryALTinterwordspacing
W.~Wang, S.~Yang, Z.~He, M.~Wang, J.~Zhang, and W.~Zhang, ``Urban perception of
  commercial activeness from satellite images and streetscapes,'' in
  \emph{Companion Proceedings of the The Web Conference 2018}, ser. WWW
  '18.\hskip 1em plus 0.5em minus 0.4em\relax Republic and Canton of Geneva,
  Switzerland: International World Wide Web Conferences Steering Committee,
  2018, pp. 647--654. [Online]. Available:
  \url{https://doi.org/10.1145/3184558.3186581}
\BIBentrySTDinterwordspacing

\bibitem{5533703}
A.~D. K.~T. {Lam} and Q.~{Li}, ``Fractal analysis and multifractal spectra for
  the images,'' in \emph{2010 International Symposium on Computer,
  Communication, Control and Automation (3CA)}, vol.~2, May 2010, pp. 530--533.

\bibitem{4767851}
J.~{Canny}, ``A computational approach to edge detection,'' \emph{IEEE
  Transactions on Pattern Analysis and Machine Intelligence}, vol. PAMI-8,
  no.~6, pp. 679--698, Nov 1986.

\bibitem{7780459}
K.~{He}, X.~{Zhang}, S.~{Ren}, and J.~{Sun}, ``Deep residual learning for image
  recognition,'' in \emph{2016 IEEE Conference on Computer Vision and Pattern
  Recognition (CVPR)}, June 2016, pp. 770--778.

\bibitem{Hartigan1979}
J.~A. Hartigan and M.~A. Wong, ``A k-means clustering algorithm,'' \emph{JSTOR:
  Applied Statistics}, vol.~28, no.~1, pp. 100--108, 1979.

\bibitem{45823}
\BIBentryALTinterwordspacing
D.~Ha, A.~Dai, and Q.~V. Le, ``Hypernetworks,'' 2017. [Online]. Available:
  \url{https://openreview.net/pdf?id=rkpACe1lx}
\BIBentrySTDinterwordspacing

\bibitem{NIPS2016_6068}
\BIBentryALTinterwordspacing
L.~Bertinetto, J.~a.~F. Henriques, J.~Valmadre, P.~Torr, and A.~Vedaldi,
  ``Learning feed-forward one-shot learners,'' in \emph{Advances in Neural
  Information Processing Systems 29}, D.~D. Lee, M.~Sugiyama, U.~V. Luxburg,
  I.~Guyon, and R.~Garnett, Eds.\hskip 1em plus 0.5em minus 0.4em\relax Curran
  Associates, Inc., 2016, pp. 523--531. [Online]. Available:
  \url{http://papers.nips.cc/paper/6068-learning-feed-forward-one-shot-learners.pdf}
\BIBentrySTDinterwordspacing

\bibitem{kingma:adam}
D.~P. Kingma and J.~Ba, ``Adam: A method for stochastic optimization,'' in
  \emph{International Conference on Learning Representations (ICLR)}, 2015.

\bibitem{Friedman00greedyfunction}
J.~H. Friedman, ``Greedy function approximation: A gradient boosting machine,''
  \emph{Annals of Statistics}, vol.~29, pp. 1189--1232, 2000.

\end{thebibliography}
\end{document}